\documentclass[10pt,twocolumn,letterpaper]{article}

\usepackage{cvpr}              %

\usepackage{color}
\usepackage{colortbl}
\usepackage[table,xcdraw,dvipsnames]{xcolor}
\usepackage{xcolor-material}
\usepackage[accsupp]{axessibility} %

\usepackage{cuted}
\usepackage{capt-of}

\usepackage{pifont}

\usepackage[normalem]{ulem}
\useunder{\uline}{\ul}{}

\newcommand{\myparagraph}[1]{\par\vspace{0.3em}\noindent\textbf{#1}\hspace{0.2em}}

\usepackage{tikz}
\usepackage{pgfplots}
\usepackage{pgfplotstable}

\pgfplotsset{compat=1.18}

\pgfplotsset{select coords between index/.style 2 args={
    x filter/.code={
        \ifnum\coordindex<#1\fi
        \ifnum\coordindex>#2\fi
    }
}}

\pgfplotsset{
    standard/.style={
        enlargelimits=true,
        enlarge x limits=false,
        ymajorgrids=true,
        grid style=dashed,
        xtick pos=left,
        ytick pos=left,
        max space between ticks=25,
        set layers=tick labels on top%
    },
    layers/tick labels on top/.define layer set=%
        {axis background,axis grid,axis ticks,axis lines,main,%
          axis tick labels,%
          axis descriptions,axis foreground}
        {/pgfplots/layers/standard}
}

\definecolor{cvprblue}{rgb}{0.21,0.49,0.74}
\usepackage[pagebackref,breaklinks,colorlinks,citecolor=cvprblue]{hyperref}
\usepackage{colortbl}

\def\paperID{15281} %
\def\confName{CVPR}
\def\confYear{2024}

\definecolor{somegray}{rgb}{0.5, 0.5, 0.5}
\newcommand{\darkgrayed}[1]{\textcolor{somegray}{#1}}
\makeatletter
\newcommand*\titleheader[1]{\gdef\@titleheader{#1}}
\AtBeginDocument{%
  \let\st@red@title\@title
  \def\@title{%
    \vskip-3em
    \bgroup\normalfont\large\centering\@titleheader\par\egroup
    \vskip1.5em\st@red@title}
}
\makeatother

\titleheader{\darkgrayed{This paper has been accepted for publication at the\\
IEEE Conference on Computer Vision and Pattern Recognition (CVPR), Seattle, 2024.
\copyright IEEE}}

\title{Mitigating Motion Blur in Neural Radiance Fields with Events and Frames}

\author{Marco Cannici and Davide Scaramuzza \\[5pt]
Robotics and Perception Group, University of Zurich, Switzerland
}

\begin{document}
\maketitle

\begin{abstract}
Neural Radiance Fields (NeRFs) have shown great potential in novel view synthesis. However, they struggle to render sharp images when the data used for training is affected by motion blur. On the other hand, event cameras excel in dynamic scenes as they measure brightness changes with microsecond resolution and are thus only marginally affected by blur. Recent methods attempt to enhance NeRF reconstructions under camera motion by fusing frames and events. However, they face challenges in recovering accurate color content or constrain the NeRF to a set of predefined camera poses, harming reconstruction quality in challenging conditions. This paper proposes a novel formulation addressing these issues by leveraging both model- and learning-based modules. We explicitly model the blur formation process, exploiting the event double integral as an additional model-based prior. Additionally, we model the event-pixel response using an end-to-end learnable response function, allowing our method to adapt to non-idealities in the real event-camera sensor. We show, on synthetic and real data, that the proposed approach outperforms existing deblur NeRFs that use only frames as well as those that combine frames and events by $+6.13$dB and $+2.48$dB, respectively. %
\end{abstract}

\noindent \textbf{Multimedial Material:} For videos, datasets and code visit \url{https://github.com/uzh-rpg/evdeblurnerf}.

\section{Introduction} \label{sec:introduction}

Neural Radiance Fields (NeRFs) \cite{mildenhall2020nerf} have completely revolutionized the field of 3D reconstruction and novel view synthesis, achieving unprecedented levels of details \cite{barron2022mip,barron2021mip,verbin2022ref}. As a result, they have quickly found applications in many subfields of computer vision and robotics, such as pose estimation and navigation \cite{sucar2021implicit,zhu2022neural,yen2021inerf}, image processing \cite{mildenhall2022nerf,huang2022hdr,ma2022deblur,wang2022nerf}, scene understanding \cite{liu2022unsupervised,xie2021fig,kundu2022panoptic}, surface reconstruction \cite{wang2021neus,azinovic2022neural,yu2022monosdf}, and many others.

\begin{figure}
    \centering
    \includegraphics[width=\linewidth]{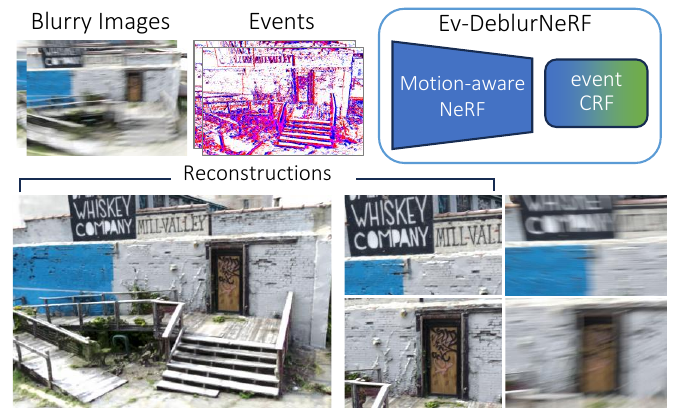}
    \caption{Ev-DeblurNeRF combines blurry images and events to recover sharp NeRFs. A motion-aware NeRF recovers camera motion and a learnable event camera response function models real camera's non-idealities, enabling high-quality reconstructions.}
    \label{fig:teaser}
\end{figure}

Leveraging multi-view consistency from calibrated images, NeRF exploits supervision from multiple view-points, enabling generalization to novel camera poses and the ability to render view-dependent color effects \cite{verbin2022ref}. However, akin to other methods relying on photometric consistency, NeRF can only deliver high-quality reconstructions when the images used for training are perfectly captured and free from any artifact. Unfortunately, perfect conditions are seldom met in the real world.

For example, in robotics, camera motion is prevalent when capturing images, often resulting in motion blur. Under such conditions, NeRFs are unable to reconstruct sharp radiance fields, thereby impeding their practical use in real-world scenes. Although recent works \cite{ma2022deblur,peng2022pdrf,wang2022bad,lee2023dp} have shown promising results in reconstructing radiance fields from motion-blurred images by learning to infer the camera motion during the exposure time, the task of recovering motion-deblurred NeRFs still remains significantly ill-posed. Existing image-based approaches typically fail when training images exhibit similar and consistent motion \cite{ma2022deblur}, and they are inherently limited by the presence of motion ambiguities and loss of texture details that cannot be recovered from blurry images alone.

In this regard, recent works have shown that event-based cameras can substantially aid the task of deblurring images captured with standard cameras \cite{zhang2023event,vitoria2023event,sun2022event,shang2021bringing}. These sensors measure brightness changes at microseconds resolution and are practically unaffected by motion blur \cite{hu2021v2e}, thus directly addressing the aforementioned ambiguities. Motivated by these advantages, the literature has recently looked into the possibility of recovering NeRFs from events \cite{hwang2023ev,rudnev2022eventnerf,klenk2022nerf,qi2023e2nerf,bhattacharya2023evdnerf}. While most of the literature \cite{hwang2023ev,rudnev2022eventnerf,bhattacharya2023evdnerf} focuses on event-only NeRFs, only two prior works \cite{qi2023e2nerf,klenk2022nerf} investigate fusing motion-blurred images with events. E-NeRF \cite{klenk2022nerf} decouples sharpness and color recovery but struggles at recovering accurate color content, as the rendered images still exhibit blurred colors around sharp edges. E$^2$NeRF \cite{qi2023e2nerf}, on the other hand, proposes to model the camera motion by combining structure from motion with an event-aided model-based deblurring process. While effective, event supervision is only applied during the exposure time, thus potentially limiting performance under challenging motion conditions.

In this work, depicted in Fig. \ref{fig:teaser}, we propose Ev-DeblurNeRF, a novel event-based deblur NeRF formulation combining learning and model-based components. Inspired by E-NeRF \cite{klenk2022nerf}, it exploits continuous event-by-event supervision to recover sharp radiance fields. But it departs from E-NeRF in that it models the blur formation process explicitely, exploiting the direct relationship between events triggered during the exposure time and the resulting blurred frames, i.e., the so-called Event Double Integral (EDI) \cite{pan2019bringing}. Unlike E$^2$NeRF \cite{qi2023e2nerf}, our approach employs this relation as additional training supervision, adding an end-to-end learnable camera response function that enables the NeRF to diverge from the model-based solution whenever inaccurate, resulting in higher-quality reconstructions. 

We validate Ev-DeblurNeRF on a novel event-based version of the Deblur-NeRF \cite{ma2022deblur} synthetic dataset, as well as on a new dataset we collected using a Color DAVIS event-based camera \cite{li2015design}. We show that Ev-DeblurNeRF recovers radiance fields that are $+6.13$dB more accurate than image-only baselines, and $+2.48$dB more accurate than NeRFs exploiting both images and events on real data. To summarize, our contributions are:
\begin{itemize}
    \item A novel approach for recovering a sharp NeRF in the presence of motion blur, incorporating both model-based priors and novel learning-based modules.
    \item A NeRF formulation that is $+2.48$dB more accurate and $6.9\times$ faster to train than previous event-based deblur NeRF methods.
    \item Two new datasets, one simulated and one collected using a Color-DAVIS346 \cite{li2015design} event camera, featuring precise ground truth poses for accurate quality assessment.
\end{itemize}

\section{Related Works} \label{sec:related_works}

\myparagraph{Neural Radiance Fields (NeRFs)}
NeRFs \cite{mildenhall2020nerf} have gained widespread attention in the research community due to their impressive performance in generating high-quality images from novel viewpoints \cite{tewari2022advances,gao2022nerf}. As a result, ongoing research is constantly broadening NeRFs range of capabilities, extending their use even under unideal settings.
Among these, recent works have tackled the problem of recovering sharp neural radiance fields from blurry images. Deblur-NeRF \cite{ma2022deblur} proposes to simultaneously learn the latent sharp radiance field and a view-dependent blurring kernel, using only blurry images as input. PDRF \cite{peng2022pdrf} further extends the approach by employing a coarse-to-fine architecture that exploits additional scene features to guide the blur estimation and speed up convergence, while DP-NeRF \cite{lee2023dp} improves the motion estimation by imposing rigid motion constraints on all pixels. An alternative approach is BAD-NeRF \cite{wang2022bad}, which directly recovers the camera trajectory within the exposure time, taking inspiration from bundle-adjusted NeRF \cite{lin2021barf}. Despite impressive results, these methods often fail in the presence of severe camera motion or when the training views share similar motion trajectories, challenging their use with in-the-wild recordings. Our approach has a similar backbone architecture but, crucially, it additionally leverages the advantages of event-based cameras to help the reconstruction of sharp NeRFs. This allows us to recover texture and fine-grained details, resulting in improved performance and higher-quality reconstructions, even in the presence of challenging motion.

\myparagraph{Event-based image deblurring}
In recent years, event-based cameras have become increasingly popular in the field of computational photography \cite{tulyakov2021time,tulyakov2022time,wu2022video,messikommer2022multi,han2020neuromorphic} due to their high dynamic range and temporal resolution. %
Several methods have been proposed to exploit the unique characteristics of event cameras for image deblurring, starting from model-based methods, such as the event-based double integral (EDI), which explicitly model the relationship between events triggered during the exposure time and the resulting blurry frame \cite{pan2019bringing,pan2019bringing}. Subsequent works build on these approaches by refining predictions with learning-based modules \cite{jiang2020learning,wang2020event} or directly learning to deblur the image by fusing events and frames \cite{haoyu2020learning,zhang2020hybrid,xu2021motion,sun2023event,sun2022event}. These networks often pair the deblurring task with that of frame interpolation \cite{haoyu2020learning,sun2023event}, or make use of attention-based modules to further improve quality \cite{sun2022event}. 

Recently, event-based cameras have also been used to recover sharp images from a fast-moving camera by leveraging an implicit NeRF model of the scene. Ev-NeRF \cite{hwang2023ev}, later improved in Robust e-NeRF \cite{low2023robust}, exploits the event generation model \cite{gallego2022survey} to recover the underlying scene brightness, while EventNeRF \cite{rudnev2022eventnerf} extends this approach by incorporating color event-cameras. Recent methods \cite{klenk2022nerf,qi2023e2nerf} have also explored combining event-based cameras with motion-blurred images. E-NeRF \cite{klenk2022nerf} shows that incorporating an event supervision loss can enhance the recovery of sharp edges, but it struggles to restore sharp colors due to the lack of explicit blur modeling. Similar to ours, E$^2$NeRF \cite{qi2023e2nerf} follows Deblur-NeRF \cite{ma2022deblur} by modeling the camera motion during the exposure time. Notably, in our approach, we exploit continuous event-by-event supervision and employ a novel learnable camera response function that better adapts to real data, resulting in improved reconstruction under fast motion.

\section{Method} \label{sec:method}
\begin{figure*}
    \centering
    \includegraphics[width=\linewidth]{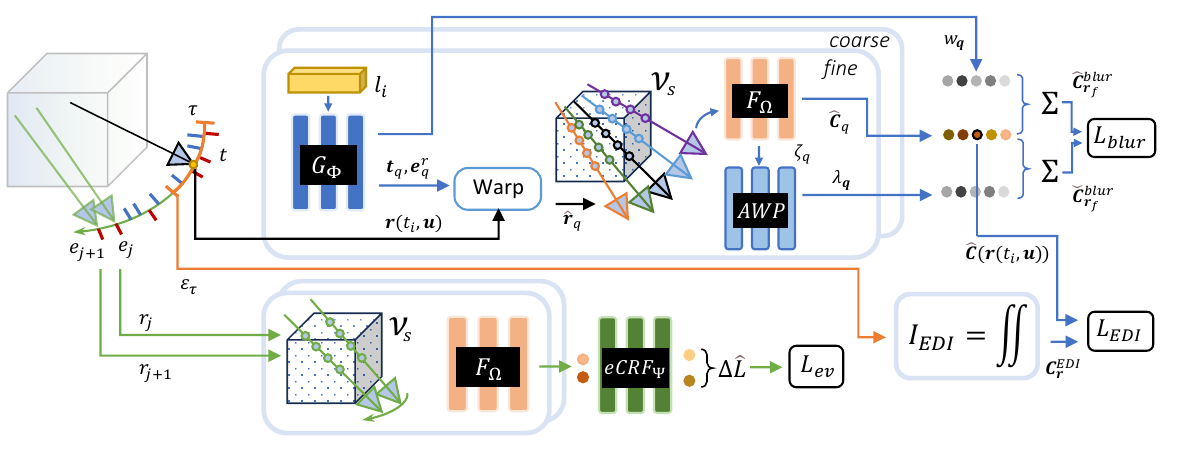}
    \caption{Architecture of the proposed Ev-DeblurNeRF model. For each given ray $\mathbf{r}(\mathbf{u},t)$, placed at the center of the exposure time $\tau$, we estimate a set of warped rays $\mathbf{r}_q$ using $G_{\boldsymbol{\Phi}}$. We then sample features from an explicit volume $\mathcal{V}$ and fed these features to $F_\Omega$ to compute blurry colors through weighted averaging with $L_\text{blur}$. We supervise the color at the center of the exposure time through $\mathcal{L}_{EDI}$ by recovering a prior-based sharp color using the event double integral, considering all events in the exposure time. Finally, we sample a pair of two consecutive events, and supervise their brightness difference, modulated by eCRF, using the observed polarity value via $\mathcal{L}_{ev}.$}
    \label{fig:network}
\end{figure*}

The proposed Ev-DeblurNeRF aims to recover a latent sharp representation of the scene given a sequence of timestamped blurry colored images $\{(\mathbf{C}^\text{blur}_i, t_i)\}_{i=1}^{N_I}$ and events $\mathcal{E}=\{\mathbf{e}_j=(\mathbf{u}_j, t_j, p_j)\}_{j=1}^{N_E}$, specifying that either an increase or decrease in brightness (as indicated by the polarity $p_j \in \{-1,1\}$) has been detected at a certain time instant $t_j$ and pixel $\mathbf{u}_j = (u_j, v_j)$. %
Our method employs recent NeRF-based deblurring modules \cite{lee2023dp,peng2022pdrf} for fast convergence and adapts them to effectively exploit event-based information. Events in our approach serve a threefold purpose: (i) as sharp brightness supervision obtained through a single integral loss \cite{gallego2022survey,klenk2022nerf,rudnev2022eventnerf,hwang2023ev}, (ii) as prior, in the form of the Double Integral (EDI) \cite{pan2019bringing}, and lastly (iii) as a learnable event-based camera response function (CRF) that enables adapting to real event-based data. Fig. \ref{fig:network} provides an overview of our proposed method. In the following section, we introduce the basics of event integrals, while in Sec. \ref{sec:evdeblurnerf} we describe the building blocks of our network.

\subsection{Preliminaries} 

\myparagraph{Event-based Single Integral.} \label{sec:preliminaries,par:esi}
Let's denote the instantaneous intensity at a monochrome pixel $\mathbf{u}$ on a given time $t$ as $I(\mathbf{u},t)$. An event $\mathbf{e}_j$ indicates that at time $t_j$, the logarithmic brightness measured at the pixel location has changed by $p_j \cdot \Theta_{p_j}$ from the last time $t_{j-1}$ an event has been generated from the same pixel location. The quantity $\Theta_{p_j} \in \mathbb{R}^+$ is a predefined threshold that controls the sensitivity to brightness changes. It follows that:
\begin{equation} \label{eq:esi_single}
    log(I(\mathbf{u},t_j)) - log(I(\mathbf{u},t_{j-1})) = \Delta L(t_{j-1}, t_j) = p_j \cdot \Theta_{p_j}.
\end{equation} 

Considering the events collected in a time period $\Delta t$ and denoting as $L(\mathbf{u}, t) = log(I(\mathbf{u},t))$ the logarithmic intensity, the following relation, here called Event-based Single Integral (ESI), holds:
\begin{equation} \label{eq:esi}
    L(t+\Delta t)-L(t)= \Theta \cdot \mathbf{E}(t) = \Theta \int_{t}^{t+\Delta t} p  \delta(\tau) \mathrm{d\tau},
\end{equation}
where we dropped the dependency from the pixel location and the polarity $p_j$ in the threshold $\Theta$ for readability, with $\delta(\tau)$ an impulse function with unit integral. Besides providing a relation between the difference in instantaneous brightness perceived at two instants and the events captured in between, Equation \eqref{eq:edi}, rewritten as $I(t+\Delta t) = I(t) \cdot \text{exp}(\Theta \cdot \mathbf{E}(t))$, also introduces a way of warping the instantaneous brightness forward or backward in time using the accumulated brightness $\Theta \cdot \mathbf{E}(t)$ measured by the event camera. This relation is utilized in the following.

\myparagraph{Event-based Double Integral.} \label{par:edi}
Let's now recall that the physical image formation process of a standard frame-based camera can be mathematically represented as integrating a sequence of latent sharp images acquired during a fixed exposure time $\tau$:
\begin{equation} \label{eq:blur}
\mathbf{I}^\text{blur}(\mathbf{u},t) = \frac{1}{\tau} \int_{t-\tau/2}^{t+\tau/2} I(\mathbf{u},h) \,\mathrm{dh},
\end{equation}
where $\mathbf{I}^\text{blur}$ is the captured image, which we consider affected by motion blur.

Following \cite{pan2019bringing}, by combining Equation \eqref{eq:edi} with \eqref{eq:blur}, we can finally draw a connection between the blurred image observed at time $t$, the events recorded during the exposure interval $\Delta T = [t-\tau/2, t+\tau/2]$ and the underlying latent sharp image $I(\mathbf{u},t)$ at time $t$:
\begin{equation} \label{eq:edi}
\mathbf{I}^\text{blur}(\mathbf{u},t) = \frac{I(\mathbf{u},t)}{\tau} \int_{t-\tau/2}^{t+\tau/2} \exp \left(\Theta \mathbf{E}(h)\right) \,\mathrm{dh}.
\end{equation}
Solving for $I(\mathbf{u},t)$, we obtain a model-based deblur of $\mathbf{I}^\text{blur}$,  guided by the events. In the following, we use this quantity as a prior to supervise our network during training.

\subsection{Event-Aided Deblur-NeRF} \label{sec:evdeblurnerf}
Our architecture takes inspiration from prior works \cite{ma2022deblur,lee2023dp,peng2022pdrf}, and is depicted in Figure \ref{fig:network}.
We aim to recover the scene as a radiance field, implemented by an MLP $F_{\Omega}$, blindly, by directly modeling the blur formation process at each exposure. Analogous to Equation \eqref{eq:blur}, a blurry color observation generated by the ray $\textbf{r}(\mathbf{u},t_i)$ cast by pixel $\mathbf{u}$ during its exposure can be described as the integral of the sharp colors observed by the ray in a time interval $\Delta T_i= [t_i-\tau/2, t_i+\tau/2]$. %

Similarly to \cite{lee2023dp}, we learn to estimate the motion of each ray implicitly using a neural module $G_{\boldsymbol{\Phi}}$. We discretize motion in a finite set of $M$ observations and learn an $SE(3)$ field that rigidly warps pixel rays to each position $q$:
\begin{equation}
    (\mathbf{e^r}_q, \mathbf{t}_q, w_q) = G_{\boldsymbol{\Phi}}(\mathcal{R}({\textbf{\textit{l}}_i}); \mathcal{T}({\textbf{\textit{l}}_i}); \mathcal{W}({\textbf{\textit{l}}_i})),
\end{equation}
where $\textbf{\textit{l}} \in \mathbb{R}^E$ is a shared learned image embedding, and $\mathcal{R}$, $\mathcal{T}$ and $\mathcal{W}$ are independent MLPs that predict, respectively, a set of rotation matrices $\mathbf{e}^r_q \in SO(3)$, translation vectors $\mathbf{t}_q \in \mathcal{R}^3$, and view weights $w_q \in \mathcal{R}$, one for each discrete position $q$. %
The warped rays can thus be finally obtained as $\hat{\mathbf{r}}_{q} = \mathbf{e^r}_q \mathbf{r}(\mathbf{u}, t_i) + \mathbf{t}_q$.

Following NeRF \cite{gao2022nerf}, we render the color at each ray by first sampling a set of 3D points along each ray, and then query a pair of MLPs, one coarse- and one fine-grained, $F_{\Omega}^c$ and $F_{\Omega}^f$, to obtain colors and density at each location. Volumetric rendering is then finally used to estimate colors $\hat{\mathbf{C}}_q$ at the predicted camera positions, which are finally fused into a blurry observation
\begin{equation} \label{eq:blur-synthesis}
    \hat{\mathbf{C}}^\text{blur}(\mathbf{r}(\mathbf{u},t_i))=g\left(\sum_{q=1}^{M-1} w_{q} \hat{\mathbf{C}}_{q}\right),
\end{equation}
where $g(\cdot)$ is a gamma correction function. 
Inspired by \cite{lee2023dp}, we further refine the composite weights using an adaptive weight proposal network $\lambda_q = \mathcal{AWP}(\zeta_{q}, \textbf{\textit{l}}_i, \mathbf{d}_{q})$, which takes the ray's samples features $\zeta_{q}$, directions $\mathbf{d}_{q}$ and image embedding $\mathbf{l}_i$ to produce refined weights. We use these refined weights in Equation \eqref{eq:blur-synthesis} in place of $w_q$ to obtain refined colors $\tilde{\mathbf{C}}^\text{blur}$.

The thus rendered synthetic blurry pixel is finally supervised with a ground truth observation $\mathbf{C}_\text{gt}$ through:
\begin{gather}
    E^b(\mathbf{C}^\text{blur}_\mathbf{r}) = \left\|\mathbf{C}^\text{blur}_\mathbf{r} -\mathbf{C}_\text{gt}^\text{blur}(\mathbf{r})\right\|_2^2 \\
    \label{eq:loss-blur}
    \mathcal{L}_\text{blur} = \hspace{-2pt}\tfrac{1}{\lvert\mathcal{R}_b\rvert}\hspace{-4pt} \sum_{\mathbf{r} \in \mathcal{R}_b} E^b(\hat{\mathbf{C}}^\text{blur}_{\mathbf{r}_c}) + E^b(\hat{\mathbf{C}}^\text{blur}_{\mathbf{r}_f}) + E^b(\tilde{\mathbf{C}}^\text{blur}_{\mathbf{r}_f}),
\end{gather}
where we consider a batch of pixels $\mathcal{R}_b$, and rewrite $\mathbf{C}^\text{blur}_\mathbf{r} = \mathbf{C}^\text{blur}(\mathbf{r})$. Subscripts $c$ and $f$ indicate if values are obtained through $F_{\Omega}^c$ or $F_{\Omega}^f$, while $\tilde{\hspace{6pt}}$ if adaptive weights are used.

\myparagraph{Event-based supervision via learned event-CRF. } %
When the scene is also captured by an event-based camera, as in our case, blur-free microsecond-level measurements can be exploited to further assist the reconstruction of a sharp radiance field, leveraging the relation in Equation \eqref{eq:esi_single} between brightness and generated events. We do so by synthesizing the left-hand side of \eqref{eq:esi_single}, i.e., the log brightness difference perceived by the event pixel, through volumetric rendering while we take the right-hand side as a ground truth supervision, given recorded event pairs. 

In particular, we estimate the log-brightness at each event $\mathbf{e}_j$, produced by the event pixel $\mathbf{u}$ at time $t_j$, as:
\begin{equation} \label{eq:ecrf}
    \hat{L}(t_j,\mathbf{u}) = log(h(eCRF_{\Psi}(\hat{\mathbf{C}}({\mathbf{r}_j}), p_j))),
\end{equation}
where we obtain $\hat{\mathbf{C}}({\mathbf{r}_j})$ via volumetric rendering \cite{mildenhall2020nerf} by rendering the ray  $\mathbf{r}_j = \mathbf{r}(\mathbf{u}, t_j)$ cast from the camera pose $\mathbf{T}(t_j) \in SE(3)$, approximated via spherical linear interpolation \cite{shoemake1985animating} of the available known camera poses. Here, $eCRF_{\Psi}$ is an MLP that produces a modulated signal $\hat{\mathbf{C}}_e \in \mathbb{R}^3$ from the rendered color $\hat{\mathbf{C}}$ and the polarity $p_j$, while $h(\cdot)$ is a luma conversion function, implemented following the BT.601 \cite{bt2011studio} standard. 

Given a pair of consecutive events at time $t_{j-1}$ and $t_j$, we first estimate the log-brightness difference as $\Delta \hat{L}(\mathbf{u},t_j) = \hat{L}(\mathbf{u},t_j) - \hat{L}(\mathbf{u},t_{j-1})$ and then compare it with that observed by the event camera, $ \Delta L$, as follows:
\begin{gather}
    E^e(\Delta \hat{L}^t_\mathbf{u}) = \left\| \Delta \hat{L}^t_\mathbf{u} - \Delta L^t_\mathbf{u}\right\|_2^2 \\
    \label{eq:loss-event}
    \hspace{-0.5em}\mathcal{L}_\text{ev} = \hspace{-1pt}\tfrac{1}{\lvert\mathcal{U}_e\rvert}\hspace{-1pt} \hspace{-0.9em}\sum_{(t,\mathbf{u}) \in \mathcal{U}_e} \hspace{-1em}E^e(\Delta \hat{L}^t_{\mathbf{u}_c}) \hspace{-0.2em}+\hspace{-0.2em} E^e(\Delta \hat{L}^t_{\mathbf{u}_f}) \hspace{-0.2em}+\hspace{-0.2em} E^e(\Delta \tilde{L}^t_{\mathbf{u}_f}),
\end{gather}
where we use the compact form $\hat{L}^t_\mathbf{u}$ for $\hat{L}(t,\mathbf{u})$, and apply the supervision on fine and coarse levels, as well as on adaptively refined colors. $\mathcal{U}_e$ selects pairs of pixels $\mathbf{u}$ and timestamps $t$ corresponding to received events.
Our experiments reveal that applying $\mathcal{L}_\text{ev}$ not only during image exposures but also between frames, similar to \cite{klenk2022nerf}, helps in viewpoints with scarce RGB coverage, as common with fast motion.

In Equation \eqref{eq:loss-event}, we assume the ideal event generation model of \eqref{eq:esi}. However, real event pixels deviate from the ideal case  \cite{hu2021v2e}. 
Our proposed event CRF function $eCRF_{\Psi}$ learns to compensate for potential mismatches between the ideal model and that of the camera at hand, filling the gap between the rendered color space and the brightness change perceived by the event sensor. Note that, when a color event camera is used, as the one in \cite{li2015design}, pixels record color intensity changes following a Bayer pattern. We remove the luma conversion $h(\cdot)$ function in Equation \eqref{eq:ecrf}, and directly apply the previous loss to the color channel each pixel is responsible for. We refer to this version of the loss as $\mathcal{L}_\text{ev-color}$.

\myparagraph{Double integral supervision. }
The eCRF just introduced provides an effective way of handling unmodeled event pixel behaviors. However, blindly recovering the event camera response to colors is not trivial since the only direct source of color supervision comes from Equation \eqref{eq:loss-blur}. In practice, the optimization problem in \eqref{eq:loss-event} is under constrained, as the loss, acting on the event CRF, is free to enhance texture details in the radiance field as long as they correctly render once blurred through Equation \eqref{eq:blur}. Inspired by recent works \cite{li2021neural}, which suggest facilitating NeRF optimization through priors, we propose here to exploit the relationship in \eqref{eq:edi} to further constrain the NeRF training. 

In particular, we consider every original ray $\mathbf{r} \in \mathcal{R}_b$ sampled when optimizing Eq. \eqref{eq:loss-blur} originating from the mid-exposure pose of image $I_i^\text{blur}$, i.e., the rays rendering the latent sharp pixels $\mathbf{C}(\mathbf{r})$. If we simplify and assume these pixels are monochrome, they correspond to $I(\mathbf{u},t_i)$ in Equation \eqref{eq:edi}. Given this observation, 
we first rewrite \eqref{eq:edi} by solving for $I(\textbf{u}, t_i)$, and then evaluate it channel-wise for the given image at time $t_i$ and ray $\mathbf{r}$, using the observed blurry color $\mathbf{C}^\text{blur}$ and the events received at pixel $\mathbf{u}$. We finally collect channels into $\mathbf{C}^{EDI}_\mathbf{r} = \left[I^R(\textbf{u}, t_i),I^G(\textbf{u}, t_i),I^B(\textbf{u}, t_i)\right]$, obtaining a model-based sharp latent color.
We use this color as a prior in:
\begin{gather}
    E^\text{EDI}(\hat{\mathbf{C}}_r) = \left\|\hat{\mathbf{C}}_\mathbf{r} -\mathbf{C}_\mathbf{r}^\text{EDI}\right\|_2^2 \\
    \label{eq:loss-edi}
    \mathcal{L}_\text{EDI} = \tfrac{1}{\lvert\mathcal{R}_b\rvert}\hspace{-2pt} \sum_{\mathbf{r} \in \mathcal{R}_b} E^\text{EDI}(\hat{\mathbf{C}}_{\mathbf{r}_c}) + E^\text{EDI}(\hat{\mathbf{C}}_{\mathbf{r}_f})
\end{gather}

\myparagraph{Fast NeRF via explicit features. } The additional event-based supervision introduced in Equation \eqref{eq:loss-event}, while enabling the reconstruction of a high-fidelity sharp NeRF, does come with a notable effect on the training time. Indeed, on top of the rays $\mathcal{R}_b$, needed for optimizing Equations \eqref{eq:loss-blur} and \eqref{eq:loss-edi}, we also consider an additional pair of rays in $\mathcal{U}_e$ which we employ to render brightness changes across time. We overcome this aspect by taking inspiration from previous works \cite{peng2022pdrf,chen2022tensorf} showing that additional explicit features can ease convergence, making the training faster.

Inspired by the hybrid design in \cite{peng2022pdrf}, we enhance the capabilities of $F_{\Omega}^c$ and $F_{\Omega}^f$ by incorporating dedicated TensoRF \cite{chen2022tensorf} volumes, which we employ as additional input feature spaces for the MLPs. In particular, given a ray $\mathbf{r}_\mathbf{u}$ and a set of coarse and fine points $\{\mathbf{x}^c_k\}_{k=1}^S$ and $\{\mathbf{x}^f_k\}_{k=1}^S$ along the ray, we first sample feature volumes:
\begin{equation} \label{eq:voxels}
\begin{aligned}
    {f_s}_k^c &= \mathcal{V}_s(\mathbf{x}^c_k), \,\,
    {f_s}_k^f = \mathcal{V}_s(\mathbf{x}^f_k),\\
    {f_l}_k^c &= \mathcal{V}_l(\mathbf{x}^c_k), \,\,
    {f_l}_k^f = \mathcal{V}_l(\mathbf{x}^f_k),
\end{aligned}    
\end{equation}
with $\mathcal{V}_s$ and $\mathcal{V}_l$, respectively, a small and a large TensoRF \cite{chen2022tensorf} volume. We use ${f_s}_k^c$ as additional features in $F_{\Omega}^c$, while we employ all the features as input to the fine-grained MLP $F_{\Theta}^f$. The structure of $F_{\Omega}^c$ and $F_{\Omega}^f$ is analogous to that of the original NeRF \cite{mildenhall2020nerf}, with the only difference that the MLP predicting $\sigma$ also takes these extra features as input. 
\section{Experiments} \label{sec:experiments}

\begin{table*}
\caption{Quantitative comparison on the synthetic Ev-DeblurBlender dataset. Best results are reported in bold.}

\label{tab:synth_results}
\resizebox{\textwidth}{!}{%
\setlength{\tabcolsep}{5.4pt}
\begin{tabular}{|c|ccc|ccc|ccc|ccc|ccc|}
\cline{2-16}
\multicolumn{1}{l|}{}  & \multicolumn{3}{c|}{\textsc{Factory}}                                 & \multicolumn{3}{c|}{\textsc{Pool}}                                   & \multicolumn{3}{c|}{\textsc{Tanabata}}                                & \multicolumn{3}{c|}{\textsc{Trolley}}                            & \multicolumn{3}{c|}{\textsc{Average}}                                  \\
\multicolumn{1}{l|}{}  & PSNR$\uparrow$ & LPIPS $\downarrow$ & SSIM$\uparrow$ & PSNR$\uparrow$ & LPIPS $\downarrow$ & SSIM$\uparrow$ & PSNR$\uparrow$ & LPIPS $\downarrow$ & SSIM$\uparrow$ & PSNR$\uparrow$ & LPIPS $\downarrow$ & SSIM$\uparrow$ & PSNR$\uparrow$ & LPIPS $\downarrow$ & SSIM$\uparrow$ \\ \hline
DeblurNeRF \cite{ma2022deblur}             & 24.52             & 0.25                 & 0.79              & 26.02            & 0.34                 & 0.69              & 21.38             & 0.28                 & 0.71              & 23.58             & 0.22                 & 0.79              & 23.87             & 0.27                 & 0.74              \\
BAD-NeRF \cite{wang2022bad}                & 21.20             & 0.22                 & 0.64              & 27.13            & 0.23                 & 0.70              & 20.89             & 0.25                 & 0.65              & 22.76             & 0.18                 & 0.73              & 22.99             & 0.22                 & 0.68           \\
PDRF \cite{peng2022pdrf}                   & 27.34             & 0.17                 & 0.87              & 27.46            & 0.32                 & 0.72              & 24.27             & 0.20                 & 0.81              & 26.09             & 0.15                 & 0.86              & 26.29             & 0.21                 & 0.81              \\
DP-NeRF \cite{lee2023dp}                & 26.77             & 0.20                 & 0.85              & 29.58            & 0.24                 & 0.79              & 27.32             & 0.11                 & 0.85              & 27.04             & 0.14                 & 0.87              & 27.68             & 0.17                 & 0.84              \\ \hline
MPRNet \cite{zamir2021multi} + NeRF          & 19.09             & 0.37                 & 0.56              & 25.49            & 0.39                 & 0.64              & 17.79             & 0.42                 & 0.51              & 19.82             & 0.31                 & 0.62              & 20.55             & 0.37                 & 0.58              \\
PVDNet \cite{son2021recurrent} + NeRF          & 22.50             & 0.29                 & 0.71              & 23.89            & 0.43                 & 0.52              & 20.26             & 0.33                 & 0.64              & 22.49             & 0.25                 & 0.74              & 22.28             & 0.32                 & 0.65              \\
EFNet \cite{sun2022event} + NeRF           & 20.91             & 0.32                 & 0.63              & 27.03            & 0.31                 & 0.73              & 20.68             & 0.31                 & 0.64              & 21.69             & 0.25                 & 0.69              & 22.58             & 0.30                 & 0.67              \\
EFNet* \cite{sun2022event} + NeRF      & 29.01             & 0.14                 & 0.87              & 29.77            & 0.18                 & 0.80              & 27.76             & 0.11                 & 0.87              & 29.40             & 0.94                 & 0.89              & 28.99             & 0.34                 & 0.86              \\ \hline
ENeRF \cite{klenk2022nerf}                  & 22.46             & 0.19                 & 0.79              & 25.51            & 0.28                 & 0.72              & 22.97             & 0.16                 & 0.83              & 21.07             & 0.20                 & 0.80              & 23.00             & 0.21                 & 0.79              \\
E$^2$NeRF \cite{qi2023e2nerf}                 & 24.90             & 0.17                 & 0.78              & 29.57            & 0.18                 & 0.78              & 23.06             & 0.19                 & 0.74              & 26.49             & 0.10                 & 0.85              & 26.00             & 0.16                 & 0.78              \\
(Ours) Ev-DeblurNeRF-\,- & \textbf{32.84}    & \textbf{0.05}                 & \textbf{0.94}     & 31.45            & \textbf{0.14}        & \textbf{0.84}     & \textbf{29.20}    & \textbf{0.06}        & \textbf{0.92}     & \textbf{30.60}    & \textbf{0.06}        & \textbf{0.93}     & \textbf{31.02}    & \textbf{0.08}        & \textbf{0.91}     \\
(Ours) Ev-DeblurNeRF          & 31.79             & 0.06                 & 0.93              & \textbf{31.51}   & \textbf{0.14}        & \textbf{0.84}     & 28.67             & 0.08                 & 0.90              & 29.72             & 0.07                 & 0.92              & 30.42             & \textbf{0.08}        & 0.90              \\ \hline
\end{tabular}%
}
\end{table*}
\begin{table*}
\caption{Quantitative comparison on the real-world Ev-DeblurCDAVIS dataset. Best results are reported in bold.}

\label{tab:real_results}
\resizebox{\textwidth}{!}{%
\setlength{\tabcolsep}{2pt}
\begin{tabular}{|c|ccc|ccc|ccc|ccc|ccc|ccc|}
\cline{2-19}
\multicolumn{1}{l|}{} & \multicolumn{3}{c|}{\textsc{Batteries}}                & \multicolumn{3}{c|}{\textsc{Power supplies}}           & \multicolumn{3}{c|}{\textsc{Lab Equipment}}            & \multicolumn{3}{c|}{\textsc{Drones}}                   & \multicolumn{3}{c|}{\textsc{Figures}}                  & \multicolumn{3}{c|}{\textsc{Average}}                   \\
\multicolumn{1}{l|}{} & PSNR$\uparrow$ & LPIPS$\downarrow$ & SSIM$\uparrow$ & PSNR$\uparrow$ & LPIPS$\downarrow$ & SSIM$\uparrow$ & PSNR$\uparrow$ & LPIPS$\downarrow$ & SSIM$\uparrow$ & PSNR$\uparrow$ & LPIPS$\downarrow$ & SSIM$\uparrow$ & PSNR$\uparrow$ & LPIPS$\downarrow$ & SSIM$\uparrow$ & PSNR$\uparrow$ & LPIPS$\downarrow$ & SSIM$\uparrow$ \\ \hline
DP-NeRF \cite{lee2023dp} + TensoRF \cite{chen2022tensorf}           & 26.64        & 0.27            & 0.81         & 25.74        & 0.32            & 0.77         & 27.49        & 0.31            & 0.80         & 26.52        & 0.30            & 0.81         & 27.76        & 0.34            & 0.77         & 26.83        & 0.31            & 0.79         \\
EDI \cite{pan2019bringing} + NeRF              & 28.66        & 0.12            & 0.87         & 28.16        & 0.09            & 0.88         & 31.45        & 0.13            & 0.89         & 29.37        & 0.10            & 0.88         & 31.44        & 0.12            & 0.88         & 29.82        & 0.11            & 0.88         \\ \hline
E$^2$NeRF                & 30.57        & 0.12            & 0.88         & 29.98        & 0.11            & 0.87         & 30.41        & 0.16            & 0.86         & 30.41        & 0.14            & 0.87         & 31.03        & 0.14            & 0.85         & 30.48        & 0.13            & 0.87         \\
(Ours) Ev-DeblurNeRF  & \textbf{33.17}        & \textbf{0.05}            & \textbf{0.92}         & \textbf{32.35}        & \textbf{0.06}            & \textbf{0.91}         & \textbf{33.01}        & \textbf{0.08}            & \textbf{0.91}         & \textbf{32.89}        & \textbf{0.05}            & \textbf{0.92}         & \textbf{33.39}        & \textbf{0.07}            & \textbf{0.90}         & \textbf{32.96}        & \textbf{0.06}            & \textbf{0.91}         \\ \hline
\end{tabular}%
}
\end{table*}

\subsection{Implementation Details. }
\myparagraph{Training.} We build our event-based architecture starting from the Pytorch implementation of DP-NeRF \cite{lee2023dp}. We use a batch size of $1024$ for rays $\mathcal{R}_b$ and $2048$ for rays $\mathcal{U}_e$, and sample $64$ coarse and additional $64$ fine points along each ray. Following \cite{peng2022pdrf}, we set the number of motion locations to $M = 9$. We use Adam \cite{kingma2014adam} to optimize the multi-objective loss $\mathcal{L} = \lambda_b \mathcal{L}_{blur} + \lambda_e \mathcal{L}_{event} + \lambda_{EDI} \mathcal{L}_{EDI}$, where we set $\lambda_b = \mathcal{L}_{EDI} = 1$, and $\lambda_e = 0.1$. We train the model for a total of $30,000$ iterations, using an initial learning rate of $5\cdot10^{-3}$, which we decrease exponentially to $5\cdot10^{-6}$ over the course of the training. Further details on the network architectures are provided in the supplementary material. 

\myparagraph{Ev-DeblurBlender dataset. } We evaluate our method on four synthetic scenes derived from the original DeblurNeRF \cite{ma2022deblur} work, namely, \emph{factory}, \emph{pool}, \emph{tanabata}, and \emph{trolley}. We exclude \emph{cozy room} from our conversion as the Blender rendering for this scene relies on an image denoising post-processing step. This step causes the rendered images to show temporally inconsistent artifacts when rendered at high FPS, thereby causing unrealistic event simulation. Differently from \cite{ma2022deblur}, where blurry images are obtained by randomly moving the camera at each pose, we use a single fast continuous motion, derived from DeblurNeRF's original poses, lasting around $1$s. We simulate a $40$ms exposure time by averaging together, in linear RGB space, images rendered at $1000$ FPS. We then use the same set of images to generate synthetic events using event simulation \cite{rebecq2018esim}, making use of a balanced $\Theta=0.2$ event threshold and monochrome events.

\myparagraph{Ev-DeblurCDAVIS dataset. } Given the lack of real-world datasets for event-based NeRF deblur that incorporate ground truth sharp reference images for quantitative assessment, we introduce a novel dataset composed of $5$ real-world scenes. We use the Color-DAVIS346 \cite{li2015design} camera for recording, which captures both color events and standard frames at $346 \times 260$ pixel resolution using a RGBG Bayer pattern. We mount the camera on a motor-controlled linear slider to capture frontal-facing scenes and use the motor encoder to obtain poses at $100$ Hz. We configure the Color-DAVIS346 with a $100$ms exposure time and collect ground truth still images first, followed by a fast motion. Scenes feature $11$ to $18$ blur training views and $5$ ground truth sharp poses with both seen and unseen views.

\myparagraph{Baselines.} We evaluate our method against frame-only methods as well as methods fusing both images and events. For the first category we follow previous works \cite{ma2022deblur,lee2023dp,lee2023dp}, and select Deblur-NeRF \cite{ma2022deblur}, BAD-NeRF \cite{wang2022bad}, DP-NeRF \cite{lee2023dp} and PDRF \cite{peng2022pdrf} as the most recent NeRF-based baselines, as well as single-image and video deblurring methods, namely MPRNet \cite{zamir2021multi} and PVDNet \cite{son2021recurrent}, followed by NeRF \cite{mildenhall2020nerf}. Similarly, for the second category, we select E-NeRF \cite{klenk2022nerf} and E$^2$-NeRF as event-based deblur NeRF architectures, and also combine frames deblurred via the events+frames EFNet \cite{sun2022event} network with NeRF \cite{mildenhall2020nerf}. We run all baselines with default hyperparameters using the official codebases. We utilize Blender poses in Ev-DeblurBlender and motor encoder poses in Ev-DeblurCDAVIS for all baselines, including E$^2$NeRF, where we compute exposure poses via spherical linear interpolation of the available ones.

\subsection{Experimental Validation}

\begin{figure*}
    \centering
    \includegraphics{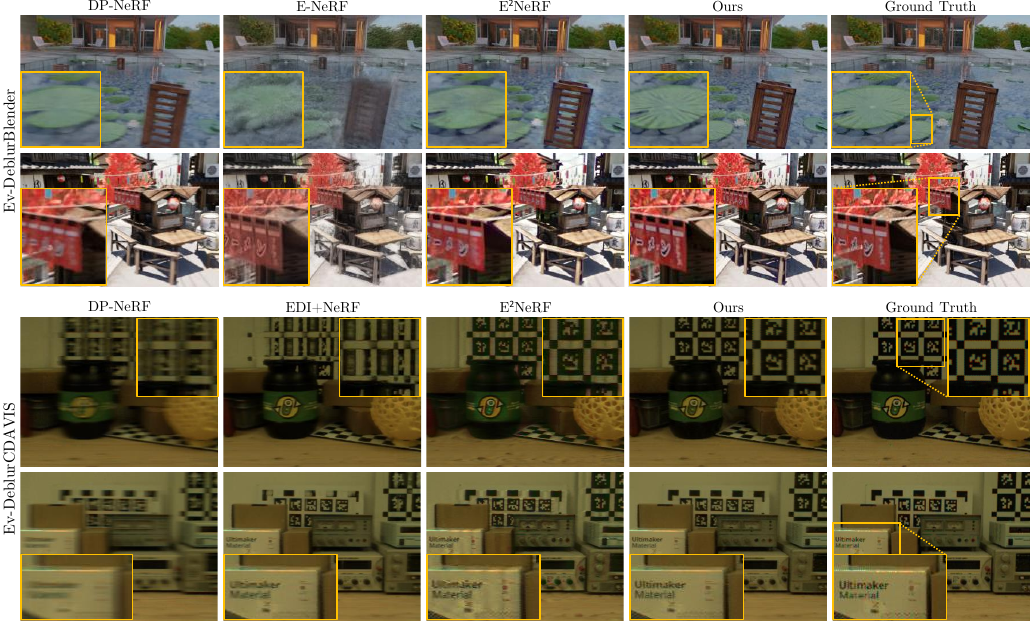}
    \caption{Qualitative comparison on synthetic (top) and real-world camera motion blur (bottom). Ev-DeblurNeRF recovers sharp and fine details, such as the letters in the last example, as well as accurate colors, outperforming other event-based and image-only methods.}
    \label{fig:qualitative_results}
\end{figure*}

\paragraph{Results on Ev-DeblurBlender. } We start the evaluation on the synthetic Ev-DeblurBlender dataset to first assess the performance of our method on an ideal case, i.e., where camera poses are accurate and the event generation model is close to the ideal case. Results are reported in Table \ref{tab:synth_results}.  We test two versions of our network. The first, which we call Ev-DeblurNeRF-\,-, does not make use of the proposed eCRF module and EDI supervision, while the second, Ev-DeblurNeRF, incorporates the complete architecture presented in Section \ref{sec:method}. We found Ev-DeblurNeRF-\,- to exhibit slightly superior performance on average on this data. As discussed in Section \ref{sec:method}, indeed, we designed the eCRF specifically to handle possible variations between RGB and events' response functions, as well as to compensate for mismatches on the event generation model. These issues are not predominant in simulated data, explaining why adding a learnable response function does not improve performance. 

Despite this, both versions largely outperform all other baselines, both event-based and frame-based. Compared to DP-NeRF \cite{lee2023dp}, which uses a similar backbone architecture, our method achieves on average a $+3.34$dB higher PSNR, a $52.9\%$ lower LPIPS \cite{zhang2018unreasonable} and $7.14\%$ higher SSIM, highlighting the improvement gained by effectively integrating event-based supervision. This is also evident when considering baselines utilizing an image-deblurring stage prior to NeRF training, which also achieve better performance when events are used. This is the case of EFNet \cite{sun2022event}, and its variant, which we name EFNet*, that we finetune on the other $3$ scenes before deblurring images of a given scene. Despite the high accuracy, these methods fail to produce scene-level consistent deblurring, causing the NeRF to reconstruct floaters and thus decreasing novel-view synthesis performance. Finally, our approach also surpasses both previous event-based deblurring NeRF methods with an average increase of $+5$dB in PSNR, a $50\%$ reduction in LPIPS, and a $16.7\%$ increase in SSIM. Notably, ENeRF, which does not explicitly model the blur formation process, struggles to recover sharp color information, while E$^2$NeRF, exclusively employing event supervision during the exposure time, fails at fully exploiting event-based data. Our method, on the contrary, overcomes both limitations, showcasing the effectiveness of the proposed approach.

\myparagraph{Results on Ev-DeblurCDAVIS. } In Table \ref{tab:real_results}, we report results obtained on data collected with a real Color-DAVIS346 camera. We select the top-performing NeRF models from the previous evaluation, namely E$^2$NeRF \cite{qi2023e2nerf} and DP-NeRF \cite{lee2023dp}, which we modify here by integrating the TensoRF modules discussed in Section \ref{sec:method} for a better comparison. Additionally, we include the performance metrics obtained by initially deblurring images using the model-based EDI deblur method, followed by NeRF. An extended analysis including all other baselines is provided in the supplementary materials. Once again, our proposed approach significantly outperforms all baselines, exhibiting an improvement of $+2.5$dB in PSNR and a $4.6\%$ increase in SSIM. A qualitative comparison, depicted in Figure \ref{fig:qualitative_results}, illustrates the capability of the proposed Ev-DeblurNeRF network in reconstructing textures and details, ultimately resulting in a higher-quality novel view synthesis. %

\begin{figure}
    \begin{tikzpicture}[every node/.append style={font=\scriptsize}]
        \def\sh{4.3cm}
        \def\w{2.5cm}
        \def\h{3.0cm}
        \begin{axis}[
            standard,
            ylabel={PSNR},
            xlabel={Training views},
            ymin=15,
            ymax=35,
            xlabel shift=-4pt,
            ylabel shift=-6pt,
            ytick={15,20,...,35},
            scale only axis,
            height=\h,
            width=\w,
            at={(0cm,0cm)},
            legend columns=5,
            legend style ={
                at={(0.0,1.0)},
                anchor=south west,
                inner sep = 2pt,
                draw = none,
                fill=none,
                /tikz/every even column/.append style={column sep=0.17cm},
                name=legendone,
            },
            legend image code/.code={
                \draw[mark repeat=2,mark phase=2]
                plot coordinates {
                (0cm,0cm)
                (0.135cm,0cm)        %
                (0.27cm,0cm)         %
                };
            },
            name=axisone
        ]  
            \pgfplotstableread[col sep=tab, colnames from=colnames]{plots/data_sparse_views.txt}\normal
            \addplot[forget plot, line width=1.05pt, color=GoogleYellow, mark=+, select coords between index={0}{3}] 
            table [x index=1, y index=2] {\normal};
            \addplot[forget plot, line width=1.05pt, color=GoogleBlue, mark=+, select coords between index={4}{7}] 
            table [x index=1, y index=2] {\normal};
            \addplot[forget plot, very thick, color=GoogleRed, mark=+, select coords between index={8}{11}] 
            table [x index=1, y index=2] {\normal};
    
            \addlegendentry{Ours};
            \addlegendimage{GoogleRed,thick}
        
            \addlegendentry{EDI + NeRF};
            \addlegendimage{GoogleBlue,thick}
        
            \addlegendentry{E$^2$NeRF};
            \addlegendimage{GoogleYellow,thick}

            \addlegendentry{PSNR};
            \addlegendimage{black,thick}
        
            \addlegendentry{LPIPS};
            \addlegendimage{black,thick,dashdotted}
        
        \end{axis}
        
        \begin{axis}[
            standard,
            ytick pos=right,
            axis y line*=right,
            axis x line=none,
            ylabel={LPIPS},
            ymin=0.05,
            ymax=0.55,
            ytick={0.05,0.175,...,0.55},
            yticklabels={0.05,0.175,0.3,0.425,0.55},
            scale only axis,
            height=\h,
            width=\w,
            ylabel shift=-9pt,
            xlabel shift=-4pt,
            at={(0cm,0cm)},
        ]
            \pgfplotstableread[col sep=tab, colnames from=colnames]{plots/data_sparse_views.txt}\normal
            \addplot[thick, dashdotted, color=GoogleYellow, mark=+, select coords between index={0}{3}] 
            table [x index=1, y index=3] {\normal};
            \addplot[thick, dashdotted, color=GoogleBlue, mark=+, select coords between index={4}{7}] 
            table [x index=1, y index=3] {\normal};
            \addplot[thick, dashdotted, color=GoogleRed, mark=+, select coords between index={8}{11}] 
            table [x index=1, y index=3] {\normal};
        \end{axis}
        
        \begin{axis}[
            standard,
            ylabel={PSNR},
            xlabel={Speed [m/s]},
            at={(\sh,0cm)},
            xlabel shift=-4pt,
            ylabel shift=-6pt,
            ymin=27,
            ymax=32,
            ytick={27,28,...,32},
            scale only axis,
            height=\h,
            width=\w,
        ]
            \pgfplotstableread[col sep=tab, colnames from=colnames]{plots/data_speed.txt}\normal
    
            \addplot[line width=1.05pt, color=GoogleYellow, mark=+, select coords between index={0}{4}] 
            table [x index=2, y index=3] {\normal};
            \addplot[line width=1.05pt, color=GoogleBlue, mark=+, select coords between index={5}{9}] 
            table [x index=2, y index=3] {\normal};
            \addplot[very thick, color=GoogleRed, mark=+, select coords between index={10}{14}] 
            table [x index=2, y index=3] {\normal};
        \end{axis}
        
        \begin{axis}[
            standard,
            ytick pos=right,
            axis y line*=right,
            axis x line=none,
            ylabel={LPIPS},
            scale only axis,
            height=\h,
            width=\w,
            at={(\sh,0cm)},
            ylabel shift=-6pt,
            xlabel shift=-4pt,
            ytick={0.1,0.15,...,0.35},
            ymin=0.1,
            ymax=0.35,
        ]
            \pgfplotstableread[col sep=tab, colnames from=colnames]{plots/data_speed.txt}\normal
            \addplot[thick, dashed, color=GoogleYellow, mark=+, select coords between index={0}{4}] 
            table [x index=2, y index=4] {\normal};
            \addplot[thick, dashed, color=GoogleBlue, mark=+, select coords between index={5}{9}] 
            table [x index=2, y index=4] {\normal};
            \addplot[thick, dashed, color=GoogleRed, mark=+, select coords between index={10}{14}] 
            table [x index=2, y index=4] {\normal};
        \end{axis}
    \end{tikzpicture}

    \caption{Analysis of the robustness to sparse training views (left) and motion blur intensity (right) on samples from the Ev-DeblurCDAVIS data.}
    \label{fig:real_analysis}
\end{figure}
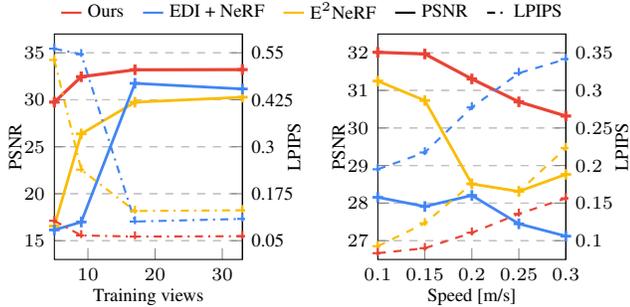

\myparagraph{Synthesis from sparse blurry views. } Utilizing the same setup used for collecting the Ev-DeblurCDAVIS dataset, we study here the robustness of the proposed approach to sparse supervision to highlight the advantage of using events not only within exposure but also in between frames. We collect an additional, longer, sequence with a back-and-forth motion and train the proposed approach with an increasing number of frames $N_f \in \{5, 9, 17, 33\}$, such that each set is a subset of the next and making sure that test poses are within training views but as furthest away as possible from them. Results are reported in Figure \ref{fig:real_analysis}. Remarkably, Ev-DeblurNeRF attains the highest performance of all methods we tested, with its performance only decreasing by $3.46$dB in PSNR when passing from $33$ to just $5$ views. In contrast, E$^2$NeRF and EDI+NeRF experience a decrease of $13.71$dB and $15$dB, respectively. These methods struggle to correctly reconstruct the radiance field from viewpoints that are only weakly supervised by blurred images. Our approach, instead, is only marginally affected. More details are provided in the supplementary material.

\myparagraph{Robustness to motion blur. } In Figure \ref{fig:real_analysis}, we analyze how the performance of the proposed approach changes as we vary the motion blur intensity. We follow the same setup as before but this time vary the slider speed from $0.1$m/s to $0.3$m/s in increments of $0.05$m/s. Notably, Ev-DeblurNeRF demonstrates superior robustness, achieving a PSNR of $32.01$dB at the highest speed. In contrast, E$^2$NeRF and EDI-NeRF achieve PSNR values of $28.77$dB and $27.12$dB, respectively. We attribute the higher performance to our choice of decoupling event supervision (Eq. \eqref{eq:loss-event}) from blur estimation (Eq. \eqref{eq:loss-blur}). In contrast to E$^2$NeRF, which fixes the poses used to render blurry images, we leave the NeRF free of optimizing the best camera views to consider for blur estimation as well as their contribution, thus achieving better robustness to different degrees of motion.

\begin{table}
\caption{Ablation study on Ev-DeblurCDAVIS.}
\label{tab:real_ablations}
\resizebox{\columnwidth}{!}{%
\setlength{\tabcolsep}{3pt}
\begin{tabular}{cccccc|ccc}
$\mathcal{V}_{c,f}$ & $\mathcal{L}_{ev}$ & $\mathcal{L}_{ev-color}$     & $\mathcal{L}_{EDI}$ & eCRF & eCRF w/ p  & PSNR$\uparrow$ & LPIPS$\downarrow$ & SSIM$\uparrow$ \\ \hline
\checkmark       &                 &              &                   &            &            & 27.55             & 0.26                 & 0.80              \\
\checkmark      & \checkmark       &              &                   &            &            & 28.24             & 0.14                 & 0.85              \\
\checkmark      & \checkmark       & \checkmark   &                   &            &            & 29.28             & 0.12                 & 0.85              \\ \hline
\checkmark      & \checkmark       & \checkmark   & \checkmark        &            &            & 32.43             & 0.10                 & 0.91              \\
\checkmark      & \checkmark       & \checkmark   &                   & \checkmark &            & 30.77             & 0.11                 & 0.86              \\ \hline
\checkmark      & \checkmark       & \checkmark   & \checkmark        & \checkmark &            & 32.90             & 0.07                 & 0.91              \\ 
\checkmark      & \checkmark       & \checkmark   & \checkmark        & \checkmark & \checkmark & \textbf{33.17}    & \textbf{0.07}                 & \textbf{0.91}    \\ \hline
                & \checkmark       & \checkmark   & \checkmark        & \checkmark & \checkmark & 33.03    & 0.08                 & 0.91    \\ \hline
\end{tabular}%
}
\end{table}
\myparagraph{Ablations. } We conclude the evaluation by studying, in Table \ref{tab:real_ablations}, the contribution of all the modules introduced in Section \ref{sec:method}, using a scene derived from the \emph{Figures} sample of Ev-DeblurCDAVIS. Adding event supervision from Equation \eqref{eq:loss-event} improves PSNR by $+0.69$dB, which is further increased by $+1.04$dB when the events' color channel is considered. Similarly, adding $\mathcal{L}_{EDI}$ in Equation \ref{eq:loss-edi} as well as the proposed eCRF module, with and without additional polarity features, also results in increased performance. Next, we study the contribution of adding the $\mathcal{L}_{EDI}$ in Equation \ref{eq:loss-edi} and the eCRF module. Performance increases in both cases, with a $+3.15$dB increase when adding $\mathcal{L}_{EDI}$ and a $+1.49$dB when adding the eCRF. The highest performance is achieved when both are combined and when the eCRF also utilizes polarity as input, with an increase of $+0.74$dB, and an overall improvement of $+5.62$dB in PSNR with respect to only using images. We finally validate the use $\mathcal{V}_{c,f}$ on the full configuration. Using explicit features guarantees faster training times without sacrificing performance. We obtain a slight boost in PSNR and LPIPS but, most notably, a $\times10.8$ speedup in training convergence. This model only takes around $3$ hours and $30$ minutes for training on an NVIDIA A100 GPU, while the same network without $\mathcal{V}_{c,f}$ takes around $38$ hours on the same hardware, as it requires more iterations at a lower learning rate. Moreover, in comparison to E$^2$NeRF, which takes around $24$ hours to train, our model is $6.9$ times faster.

\myparagraph{Limitations. } We structure the proposed Ev-DeblurNeRF assuming that events and frames can be recorded from the same image sensor. While this is possible with the suggested hardware, namely a ColorDAVIS camera, not all event cameras feature both modalities. While the proposed $\mathcal{L}_{EDI}$ loss requires pixel alignment to work effectively, we believe the proposed method could still be applied in more advanced stereo setups, such as the ones in \cite{messikommer2022multi,tulyakov2022time}, especially exploiting the proposed eCRF to compensate for different sensor responses. Moreover, our method, similar to \cite{klenk2022nerf}, estimates event camera poses via interpolation of available ones. This could lead to a performance decrease in case estimated poses are far from actual ones or they are provided at a low frequency. However, we believe refinement of camera poses through event-based methods \cite{messikommer2023data,vidal2018ultimate}, or a modified approach that only computes $\mathcal{L}_{event}$ at known camera views, could help in mitigating this issue.
\section{Conclusions} \label{sec:conclusions}
We present Ev-DeblurNeRF, a novel deblur NeRF architecture that fuses blurry frames with events for sharp NeRF recovery. Our method, exploiting explicit features for fast training convergence, integrates a learnable event-based camera response function and ad-hoc event-based supervision that facilitates fine-grained details recovery. Ev-DeblurNeRF, despite being supervised by model-based priors, can adapt to non-idealities in the camera response, potentially departing from the model-based solution. We validate our method on both synthetic and real data, achieving an increase of $+4.42$dB and $+2.48$dB in PSNR,  respectively, when compared to the previous best-performing event-based baseline, and an increase of $+2.74$dB and $+6.13$dB when compared to the top-performing image-only baseline. 
\section{Acknowledgements.}
This work was supported by the National Centre of Competence in Research (NCCR) Robotics (grant agreement No. 51NF40-185543) through the Swiss National Science Foundation (SNSF), and the European Research Council (ERC) under grant agreement No. 864042 (AGILEFLIGHT).
\appendix

\begin{table*}[h!]
\captionof{table}{Extended quantitative comparison on the real-world Ev-DeblurCDAVIS dataset. Best results are reported in bold.}
\label{tab:suppl_real_results}
\resizebox{\textwidth}{!}{%
\begin{tabular}{c|ccc|ccc|ccc|ccc|ccc|ccc|}
\cline{2-19}
\multicolumn{1}{l|}{}                      & \multicolumn{3}{c|}{\textsc{Batteries}}             & \multicolumn{3}{c|}{\textsc{Power supplies}}        & \multicolumn{3}{c|}{\textsc{Lab Equipment}}         & \multicolumn{3}{c|}{\textsc{Drones}}                & \multicolumn{3}{c|}{\textsc{Figures}}               & \multicolumn{3}{c|}{\textsc{Average}}               \\
\multicolumn{1}{l|}{}                      & PSNR$\uparrow$ & LPIPS$\downarrow$ & SSIM$\uparrow$ & PSNR$\uparrow$ & LPIPS$\downarrow$ & SSIM$\uparrow$ & PSNR$\uparrow$ & LPIPS$\downarrow$ & SSIM$\uparrow$ & PSNR$\uparrow$ & LPIPS$\downarrow$ & SSIM$\uparrow$ & PSNR$\uparrow$ & LPIPS$\downarrow$ & SSIM$\uparrow$ & PSNR$\uparrow$ & LPIPS$\downarrow$ & SSIM$\uparrow$ \\ \hline
\multicolumn{1}{|c|}{BAD-NeRF* \cite{wang2022bad}}                                  & 27.32          & 0.26              & 0.82           & 26.42          & 0.32              & 0.79           & 27.84          & 0.31              & 0.81           & 26.96          & 0.31              & 0.81           & 28.21          & 0.35              & 0.77           & 27,.5	       & 0.31	 
      & 0.80           \\
\multicolumn{1}{|c|}{DP-NeRF \cite{lee2023dp} + TensoRF \cite{chen2022tensorf}}     & 26.64          & 0.27              & 0.81           & 25.74          & 0.32              & 0.77           & 27.49          & 0.31              & 0.80           & 26.52          & 0.30              & 0.81           & 27.76          & 0.34              & 0.77           & 26.83          & 0.31              & 0.79           \\
\multicolumn{1}{|c|}{PDRF \cite{peng2022pdrf}}                 & 26.82          & 0.25              & 0.81           & 25.79          & 0.31              & 0.77           & 27.70          & 0.31              & 0.81           & 26.72          & 0.29              & 0.81           & 27.80          & 0.33              & 0.77           & 26.96          & 0.30              & 0.79           \\ \hline
\multicolumn{1}{|c|}{MPRNet \cite{zamir2021multi} + NeRF}        & 27.99          & 0.21              & 0.83           & 26.89          & 0.23              & 0.78           & 27.20          & 0.28              & 0.80           & 26.98          & 0.23              & 0.80           & 28.51          & 0.29              & 0.79           & 27.52          & 0.25              & 0.80           \\
\multicolumn{1}{|c|}{PVDNet \cite{son2021recurrent} + NeRF}        & 24.65          & 0.30              & 0.72           & 23.50          & 0.30              & 0.66           & 25.04          & 0.32              & 0.72           & 24.21          & 0.31              & 0.69           & 25.92          & 0.33              & 0.72           & 24.66          & 0.31              & 0.70           \\
\multicolumn{1}{|c|}{EFNet \cite{sun2022event} + NeRF}         & 29.85          & 0.13              & 0.88           & 29.10          & 0.13              & 0.87           & 30.28          & 0.18              & 0.88           & 29.72          & 0.14              & 0.88           & 30.62          & 0.17              & 0.85           & 29.91          & 0.15              & 0.87           \\
\multicolumn{1}{|c|}{EDI \cite{pan2019bringing} + NeRF}             & 28.66          & 0.12              & 0.87           & 28.16          & 0.09              & 0.88           & 31.45          & 0.13              & 0.89           & 29.37          & 0.10              & 0.88           & 31.44          & 0.12              & 0.88           & 29.82          & 0.11              & 0.88           \\ \hline
\multicolumn{1}{|c|}{ENeRF \cite{klenk2022nerf}}                & 27.85          & 0.26          & 0.73          & 27.91          & 0.21          & 0.76          & 27.79          & 0.25          & 0.73          & 28.28          & 0.25          & 0.77          & 29.05          & 0.18          & 0.77          & 28.17          & 0.23          & 0.75        \\
\multicolumn{1}{|c|}{E$^2$NeRF \cite{qi2023e2nerf}}               & 30.57          & 0.12              & 0.88           & 29.98          & 0.11              & 0.87           & 30.41          & 0.16              & 0.86           & 30.41          & 0.14              & 0.87           & 31.03          & 0.14              & 0.85           & 30.48          & 0.13              & 0.87           \\
\multicolumn{1}{|c|}{(Ours) Ev-DeblurNeRF} & \textbf{33.17} & \textbf{0.05}     & \textbf{0.92}  & \textbf{32.35} & \textbf{0.06}     & \textbf{0.91}  & \textbf{33.01} & \textbf{0.08}     & \textbf{0.91}  & \textbf{32.89} & \textbf{0.05}     & \textbf{0.92}  & \textbf{33.39} & \textbf{0.07}     & \textbf{0.90}  & \textbf{32.96} & \textbf{0.06}     & \textbf{0.91}  \\ \hline
\end{tabular}%
}
\end{table*}
\begin{figure*}
    \centering
    \begin{minipage}{0.197\linewidth}
        \includegraphics[width=\linewidth]{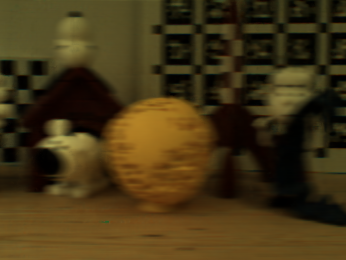}
        \resizebox{\linewidth}{!}{%
        \setlength{\tabcolsep}{2.5pt}
        \begin{tabular}{|ccccc|c|}
            $\mathcal{V}_{s,l}$ & $\mathcal{L}_\text{ev-col}$ & $\mathcal{L}_\text{EDI}$ & $eCRF$ & $eCRF$ w/p & PSNR \\
            \hline
            \checkmark & -- & -- & -- & -- & 27.55 \\
        \end{tabular}
        }
    \end{minipage}%
    \hfill
    \begin{minipage}{0.197\linewidth}
        \includegraphics[width=\linewidth]{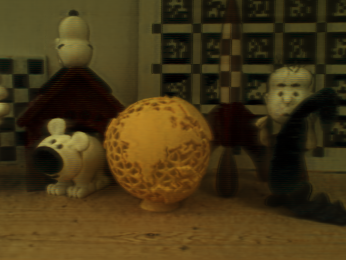}
        \resizebox{\linewidth}{!}{%
        \setlength{\tabcolsep}{2.5pt}
        \begin{tabular}{|ccccc|c|}
            $\mathcal{V}_{s,l}$ & $\mathcal{L}_\text{ev-col}$ & $\mathcal{L}_\text{EDI}$ & $eCRF$ & $eCRF$ w/p & PSNR \\
            \hline
            \checkmark & \checkmark & -- & -- & -- & 29.28\\
        \end{tabular}
        }
    \end{minipage}%
    \hfill
    \begin{minipage}{0.197\linewidth}
        \includegraphics[width=\linewidth]{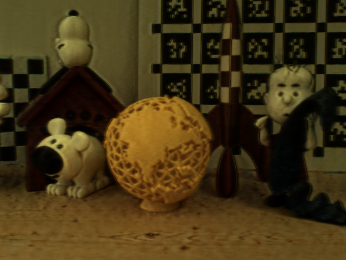}
        \resizebox{\linewidth}{!}{%
        \setlength{\tabcolsep}{2.5pt}
        \begin{tabular}{|ccccc|c|}
            $\mathcal{V}_{s,l}$ & $\mathcal{L}_\text{ev-col}$ & $\mathcal{L}_\text{EDI}$ & $eCRF$ & $eCRF$ w/p & PSNR \\
            \hline
            \checkmark & \checkmark & -- & \checkmark & -- & 30.77\\
        \end{tabular}
        }
    \end{minipage}%
    \hfill
    \begin{minipage}{0.197\linewidth}
        \includegraphics[width=\linewidth]{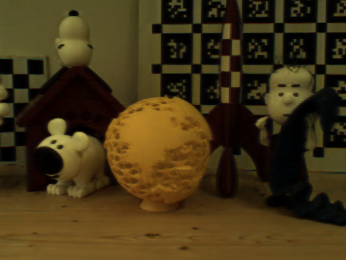}
        \resizebox{\linewidth}{!}{%
        \setlength{\tabcolsep}{2.5pt}
        \begin{tabular}{|ccccc|c|}
            $\mathcal{V}_{s,l}$ & $\mathcal{L}_\text{ev-col}$ & $\mathcal{L}_\text{EDI}$ & $eCRF$ & $eCRF$ w/p & PSNR \\
            \hline
            \checkmark & \checkmark & \checkmark & \checkmark & \checkmark & 33.17 \\
        \end{tabular}
        }
    \end{minipage}%
    \hfill
    \begin{minipage}{0.197\linewidth}
        \includegraphics[width=\linewidth]{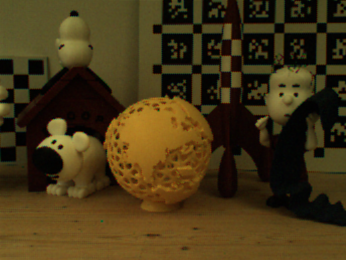}
        \begin{center}
        \vspace{-7pt}
        {\scriptsize Ground Truth}
        \vspace{3pt}
        \end{center}
    \end{minipage}
    \captionof{figure}{Qualitative ablation study of the main components of the proposed Ev-DeblurNeRF network. Tables below each picture are drawn from Table 3 of the paper, and report the configuration used and the PSNR metric achieved in each case.}
    \label{fig:qualitative_ablation}
\end{figure*}

\section{Implementation Details} \label{sec:extra_impl_details}

\paragraph{Training.} We implement Ev-DeblurNeRF building upon the DP-NeRF \cite{lee2023dp} official codebase implemented in PyTorch \cite{paszke2019pytorch}, and incorporating additional features from PDRF \cite{peng2022pdrf} and TensoRF \cite{chen2022tensorf}. We train both our Ev-DeblurNeRF network and the baselines on full-resolution images using either an NVIDIA V100, an NVIDIA RTX A6000, or an NVIDIA A100 GPU. In particular, we use $600 \times 400$ images for Ev-DeblurBlender and $346 \times 260$ for Ev-DeblurCDAVIS. Similar to \cite{ma2022deblur,lee2023dp,peng2022pdrf}, we warm up the training for the first $1,200$ iterations, by using at first only the $\mathcal{L}_{EDI}$ and $\mathcal{L}_\text{ev}$ losses and without utilizing the $eCRF$ module. Subsequently, we introduce $\mathcal{L}_{blur}$, along with the proposed $eCRF$, which we initialize as the identity function, and the blur estimation module $G_{\Phi}$, keeping the $\lambda$ parameters ($\lambda_b = \lambda_{EDI} = 1$, and $\lambda_e$ = 0.1) fixed for the entire duration of the training. To implement $L_{EDI}$, we pre-compute $C_{EDI}^\mathbf{r}$ images using Eq. (4) and directly sample them during training. When using the $\mathcal{L}_\text{ev-color}$ loss, we weigh the events' contributions by $0.4$, $0.2$, or $0.4$ depending on whether the event corresponds to a red, green, or blue channel, as green pixels appear twice as often in an RGBG Bayer pattern. We use symmetric constant thresholds for the events, setting $\Theta=0.2$ for synthetic events, and $\Theta=0.25$ when using a real camera. 

\begin{figure*}[h!]
    \begin{minipage}[b]{0.49\textwidth}
        \vspace{1pt}
        \hspace*{-6pt}\begin{tikzpicture}[every node/.append style={font=\scriptsize}]
\def\sh{2.85cm}
\def\sk{5.8cm}
\def\w{1.95cm}
\def\h{2.7cm}
    \begin{axis}[
        standard,
        ylabel={PSNR},
        xlabel={Training views},
        ymax=34,
        xlabel shift=-4pt,
        ylabel shift=-6pt,
        ytick={20,24,...,36},
        scale only axis,
        height=\h,
        width=\w,
        at={(0cm,0cm)},
        legend columns=5,
        legend style ={
        at={(-0.3,1.0)},
        anchor=south west,
        inner sep = 2pt,
         draw = none,
         fill=none,
         /tikz/every even column/.append style={column sep=0.17cm},
        name=legendone,
    },
    legend image code/.code={
    \draw[mark repeat=2,mark phase=2]
    plot coordinates {
    (0cm,0cm)
    (0.135cm,0cm)        %
    (0.27cm,0cm)         %
    };
    },
        name=axisone
    ]  
        \addlegendentry{Ours};
        \addlegendimage{GoogleRed,thick}

        \addlegendentry{E$^2$NeRF loss};
        \addlegendimage{GoogleBlue,thick}
    
        \addlegendentry{Ev-window@100Hz};
        \addlegendimage{GoogleYellow,thick}
    
        \addlegendentry{Ev-window@10-40Hz};
        \addlegendimage{GoogleGreen,thick}

        \pgfplotstableread[col sep=tab, colnames from=colnames]{plots/data_eventaccum.txt}\normal
        \addplot[forget plot, line width=1.05pt, color=GoogleBlue, mark=+, select coords between index={0}{3}] 
        table [x index=1, y index=2] {\normal};
        \addplot[forget plot, line width=1.05pt, color=GoogleYellow, mark=+, select coords between index={4}{7}] 
        table [x index=1, y index=2] {\normal};
        \addplot[forget plot, line width=1.05pt, color=GoogleGreen, mark=+, select coords between index={8}{11}] 
        table [x index=1, y index=2] {\normal};
        \addplot[forget plot, very thick, color=GoogleRed, mark=+, select coords between index={12}{15}] 
        table [x index=1, y index=2] {\normal};

    \end{axis}

    \begin{axis}[
        standard,
        ylabel={LPIPS},
        xlabel={Training views},
        at={(\sh,0cm)},
        xlabel shift=-4pt,
        ylabel shift=-6pt,
        ymax=0.4,
        scale only axis,
        height=\h,
        width=\w,
    ]
        \pgfplotstableread[col sep=tab, colnames from=colnames]{plots/data_eventaccum.txt}\normal
        \addplot[forget plot, line width=1.05pt, color=GoogleBlue, mark=+, select coords between index={0}{3}] 
        table [x index=1, y index=3] {\normal};
        \addplot[forget plot, line width=1.05pt, color=GoogleYellow, mark=+, select coords between index={4}{7}] 
        table [x index=1, y index=3] {\normal};
        \addplot[forget plot, line width=1.05pt, color=GoogleGreen, mark=+, select coords between index={8}{11}] 
        table [x index=1, y index=3] {\normal};
        \addplot[forget plot, very thick, color=GoogleRed, mark=+, select coords between index={12}{15}] 
        table [x index=1, y index=3] {\normal};
        
    \end{axis}

    \begin{axis}[
        standard,
        ylabel={SSIM},
        xlabel={Training views},
        at={(\sk,0cm)},
        xlabel shift=-4pt,
        ylabel shift=-6pt,
        scale only axis,
        height=\h,
        width=\w,
    ]
        \pgfplotstableread[col sep=tab, colnames from=colnames]{plots/data_eventaccum.txt}\normal
        \addplot[forget plot, line width=1.05pt, color=GoogleBlue, mark=+, select coords between index={0}{3}] 
        table [x index=1, y index=4] {\normal};
        \addplot[forget plot, line width=1.05pt, color=GoogleYellow, mark=+, select coords between index={4}{7}] 
        table [x index=1, y index=4] {\normal};
        \addplot[forget plot, line width=1.05pt, color=GoogleGreen, mark=+, select coords between index={8}{11}] 
        table [x index=1, y index=4] {\normal};
        \addplot[forget plot, very thick, color=GoogleRed, mark=+, select coords between index={12}{15}] 
        table [x index=1, y index=4] {\normal};
        
    \end{axis}

\end{tikzpicture}
        \caption{Analysis on event-by-event vs. event-batch losses.}
        \label{fig:real_evbyev_vs_evbatch}
    \end{minipage}
    \hspace{13pt}
    \begin{minipage}[b]{0.49\textwidth}
        \hspace*{-6pt}\begin{tikzpicture}[every node/.append style={font=\scriptsize}]
\def\sh{2.85cm}
\def\sk{5.8cm}
\def\w{1.95cm}
\def\h{2.7cm}
    \begin{axis}[
        standard,
        ylabel={PSNR},
        xlabel={Threshold $\Theta$},
        ymax=34,
        xlabel shift=-4pt,
        ylabel shift=-6pt,
        scale only axis,
        height=\h,
        width=\w,
        at={(0cm,0cm)},
        legend columns=5,
        legend style ={
        at={(0.4,1.0)},
        anchor=south west,
        inner sep = 2pt,
         draw = none,
         fill=none,
         /tikz/every even column/.append style={column sep=0.17cm},
        name=legendone,
    },
    legend image code/.code={
    \draw[mark repeat=2,mark phase=2]
    plot coordinates {
    (0cm,0cm)
    (0.135cm,0cm)        %
    (0.27cm,0cm)         %
    };
    },
        name=axisone
    ]  
        \addlegendentry{Ours};
        \addlegendimage{GoogleRed,thick}
    
        \addlegendentry{Ours w/o eCRF};
        \addlegendimage{GoogleYellow,thick}
    
        \addlegendentry{Ours w/o eCRF w/o EDI};
        \addlegendimage{GoogleBlue,thick}
    
        \pgfplotstableread[col sep=tab, colnames from=colnames]{plots/data_c.txt}\normal
        \addplot[forget plot, line width=1.05pt, color=GoogleYellow, mark=+, select coords between index={0}{3}] 
        table [x index=1, y index=2] {\normal};
        \addplot[forget plot, very thick, color=GoogleRed, mark=+, select coords between index={4}{7}] 
        table [x index=1, y index=2] {\normal};
        \addplot[forget plot, line width=1.05pt, color=GoogleBlue, mark=+, select coords between index={8}{11}] 
        table [x index=1, y index=2] {\normal};

        \draw [dotted, thin] (0.25,25) -- (0.25,35);

    \end{axis}

    \begin{axis}[
        standard,
        ylabel={LPIPS},
        xlabel={Threshold $\Theta$},
        at={(\sh,0cm)},
        xlabel shift=-4pt,
        ylabel shift=-6pt,
        ymax=0.32,
        scale only axis,
        height=\h,
        width=\w,
    ]
        \pgfplotstableread[col sep=tab, colnames from=colnames]{plots/data_c.txt}\normal
        \addplot[forget plot, line width=1.05pt, color=GoogleYellow, mark=+, select coords between index={0}{3}] 
        table [x index=1, y index=3] {\normal};
        \addplot[forget plot, line width=1.05pt, color=GoogleRed, mark=+, select coords between index={4}{7}] 
        table [x index=1, y index=3] {\normal};
        \addplot[forget plot, very thick, color=GoogleBlue, mark=+, select coords between index={8}{11}] 
        table [x index=1, y index=3] {\normal};

        \draw [dotted, thin] (0.25,0.05) -- (0.25,0.4);
        
    \end{axis}

        \begin{axis}[
        standard,
        ylabel={SSIM},
        xlabel={Threshold $\Theta$},
        at={(\sk,0cm)},
        xlabel shift=-4pt,
        ylabel shift=-6pt,
        scale only axis,
        height=\h,
        width=\w,
    ]
        \pgfplotstableread[col sep=tab, colnames from=colnames]{plots/data_c.txt}\normal
        \addplot[forget plot, line width=1.05pt, color=GoogleYellow, mark=+, select coords between index={0}{3}] 
        table [x index=1, y index=4] {\normal};
        \addplot[forget plot, line width=1.05pt, color=GoogleRed, mark=+, select coords between index={4}{7}] 
        table [x index=1, y index=4] {\normal};
        \addplot[forget plot, very thick, color=GoogleBlue, mark=+, select coords between index={8}{11}] 
        table [x index=1, y index=4] {\normal};

        \draw [dotted, thin] (0.25,0.7) -- (0.25,1.0);
        
    \end{axis}

\end{tikzpicture}
        \caption{Analysis on the robustness to model mismatches.}
        \label{fig:real_threshold}
    \end{minipage}    
    \vspace{-4pt}
\end{figure*}
\begin{table*}
\caption{Extended study on motor encoder's vs. COLMAP's poses on Ev-DeblurCDAVIS. Best results in bold, second-best underlined.}
\label{tab:suppl_real_colmap}

\centering
\setlength{\tabcolsep}{3pt}
\resizebox{\textwidth}{!}{%
\begin{tabular}{ccc|ccc|ccc|ccc|ccc|ccc|}
\cline{2-18}
\multicolumn{1}{c|}{}           & Train & Test-time  & \multicolumn{3}{c|}{\textsc{Batteries}}                      & \multicolumn{3}{c|}{\textsc{Power Supplies}}                 & \multicolumn{3}{c|}{\textsc{Lab Equipment}}                  & \multicolumn{3}{c|}{\textsc{Drones}}                         & \multicolumn{3}{c|}{\textsc{Figures}}                        \\
\multicolumn{1}{c|}{}           & poses    & refine     & PSNR$\uparrow$ & LPIPS$\downarrow$ & SSIM$\uparrow$ & PSNR$\uparrow$ & LPIPS$\downarrow$ & SSIM$\uparrow$ & PSNR$\uparrow$ & LPIPS$\downarrow$ & SSIM$\uparrow$ & PSNR$\uparrow$ & LPIPS$\downarrow$ & SSIM$\uparrow$ & PSNR$\uparrow$ & LPIPS$\downarrow$ & SSIM$\uparrow$ \\ \hline
\multicolumn{1}{|c|}{Ours}      & Motor    & --         & {\ul 33.17}    & \textbf{0.05}     & {\ul 0.92}     & \textbf{32.35} & \textbf{0.06}     & \textbf{0.91}  & {\ul 33.01}    & \textbf{0.08}     & \textbf{0.91}  & \textbf{32.89} & 0.52              & \textbf{0.92}  & 33.39          & {\ul 0.07}        & {\ul 0.90}     \\
\multicolumn{1}{|c|}{Ours}      & Motor    & \checkmark & 33.10          & \textbf{0.05}     & {\ul 0.92}     & {\ul 32.31}    & \textbf{0.06}     & \textbf{0.91}  & \textbf{33.05} & \textbf{0.08}     & \textbf{0.91}  & {\ul 32.77}    & \textbf{0.05}     & \textbf{0.92}  & {\ul 33.58}    & 0.08              & {\ul 0.90}     \\
\multicolumn{1}{|c|}{Ours}      & COLMAP   & \checkmark & \textbf{33.43} & \textbf{0.05}     & \textbf{0.93}  & 32.18          & \textbf{0.06}     & \textbf{0.91}  & {\ul 33.01}    & \textbf{0.08}     & \textbf{0.91}  & 32.69          & \textbf{0.05}     & {\ul 0.91}     & \textbf{33.88} & \textbf{0.06}     & \textbf{0.91}  \\ \hline
\end{tabular}%
}
\end{table*}

\myparagraph{Architecture.} The motion estimation module $G_{\Phi}$ is implemented following DP-NeRF \cite{lee2023dp} hyperparameters' choice, and using $M = 9$ exposure poses. Differently from \cite{lee2023dp}, we implement the image embedding $\textbf{\textit{l}}_i$ using a simple set of learnable $32$-dimensional parameters, instead of predicting them through an additional $4$-layers MLP. We found this design to be easier to optimize and yield overall better results. We follow \cite{lee2023dp} to implement the refinement $AWP$ module and employ the coarse-to-fine scheduling strategy to weight $\hat{\mathbf{C}}_{\mathbf{r}_f}^\text{blur}$ and $\tilde{\mathbf{C}}_{\mathbf{r}_f}^\text{blur}$ in $\mathcal{L}_\text{blur}$. However, we weigh their contribution equally in $\mathcal{L}_\text{ev}$ through the whole training, as we found the coarse-to-fine scheduling strategy not to improve the results. We implement $F_\Omega^c$ as a $2$-layers MLP with ReLU activation, hidden dimension $64$, and output dimension $16$, followed by a $3$-layers MLP with the same activation and hidden dimension, but output dimension $3$. We use one of the output channels of the first MLP as the predicted density, while the rest is used by the second MLP to predict colors. The structure of $F_\Omega^f$ is analogous, but we use an output dimension of $128$ for the first MLP and a $256$ hidden dimension for both MLPs. We implement $\mathcal{V}_s$ and $\mathcal{V}_l$ with vector-matrix decomposition \cite{chen2022tensorf}, using $16.7$ million voxels in $\mathcal{V}_s$ and $134.2$ million voxels in $\mathcal{V}_l$, and setting to $\{64, 16, 16\}$ the channel dimensions of the decomposed $\{X,Y,Z\}$ axes in both $\mathcal{V}_s$ and $\mathcal{V}_l$. The proposed Ev-DeblurNeRF architecture trains in around $3$ hours and $30$ minutes on an NVIDIA A100 GPU. 

\section{Extended Analysis on Ev-DeblurCDAVIS} \label{sec:more_real_results}

\paragraph{State-of-the-art comparison.} Section 4.2 of the paper provides an analysis on the Ev-DeblurCDAVIS dataset focused on the top-performing architectures selected from the synthetic evaluation. For completeness, we report in Table \ref{tab:suppl_real_results} of this supplementary material a comprehensive evaluation against all other baselines used in the paper. The trend follows that of the synthetic analysis, with image-only baselines performing worse than networks making use of either images deblurred through events or event-enhanced NeRFs. Notice that, we could not finetune EFNet \cite{sun2022event} on Ev-DeblurCDAVIS as, differently from simulation, we do not have corresponding sharp images for each blurry training view. We designed the Ev-DeblurCDAVIS dataset in such a way as to ensure reliable ground truth collection, but also to showcase the ability of our network to tackle a known limitation of image-only DeblurNeRF-like architectures. While these networks work particularly well on random motion patterns, they fail in the presence of consistent blur, i.e., when the motion pattern is similar in each exposure. This is the case of Ev-DeblurCDAVIS, where image-only baselines such as DP-NeRF \cite{lee2023dp} and PDRF \cite{peng2022pdrf} struggle to remove blur (see Figures \ref{fig:qualitative_ablation} and \ref{fig:extended_qualitative} of this supplementary material and Figure 3 of the paper). For similar reasons, BAD-NeRF diverges after a few training iterations on this dataset. We address this by fixing the
rotation matrix to ground truth and optimizing the translation vector only (reported as BAD-NeRF* in the table). Despite this, our method still significantly outperforms BAD-NeRF. Our architecture, indeed, eliminates ambiguities in motion estimation as it leverages additional event-based supervision to further constrain the NeRF recovery, resulting in significantly higher performance.

\begin{figure*}[h!]
  \begin{minipage}{0.3\textwidth}
    \centering
    \hspace*{-2pt}
\begin{tikzpicture}[every node/.append style={font=\scriptsize}]
\def\sh{0cm}
\def\w{2.8cm}
\def\h{3.5cm}
    \begin{axis}[
        standard,
        ylabel={PSNR},
        xlabel={Exposure [ms]},
        at={(\sh,0cm)},
        xlabel shift=-4pt,
        ylabel shift=-6pt,
        ymin=27,
        ymax=33,
        ytick={27,28.5,...,33},
        scale only axis,
        height=\h,
        width=\w,
        legend columns=3,
        legend style ={
            at={(-0.2,1.0)},
            anchor=south west,
            inner sep = 2pt,
            draw = none,
            fill=none,
            /tikz/every even column/.append style={column sep=0.1cm},
            name=legendone,
        },
        legend image code/.code={
            \draw[mark repeat=2,mark phase=2]
            plot coordinates {
            (0cm,0cm)
            (0.1cm,0cm)        %
            (0.2cm,0cm)         %
            };
        },
        name=axisone
    ]
        \addlegendentry{Ours};
        \addlegendimage{GoogleRed,thick}
    
        \addlegendentry{DP-NeRF};
        \addlegendimage{GoogleBlue,thick}
    
        \addlegendentry{PDRF};
        \addlegendimage{GoogleYellow,thick}

        \addlegendentry{PSNR};
        \addlegendimage{black,thick}
    
        \addlegendentry{LPIPS};
        \addlegendimage{black,thick,dashdotted}
    
        \pgfplotstableread[col sep=tab, colnames from=colnames]{plots/data_speed_synt.txt}\normal

        \addplot[line width=1.05pt, color=GoogleYellow, mark=+, select coords between index={0}{4}] 
        table [x index=1, y index=2] {\normal};
        \addplot[line width=1.05pt, color=GoogleBlue, mark=+, select coords between index={5}{9}] 
        table [x index=1, y index=2] {\normal};
        \addplot[very thick, color=GoogleRed, mark=+, select coords between index={10}{14}] 
        table [x index=1, y index=2] {\normal};        
    \end{axis}

    \begin{axis}[
        standard,
        ytick pos=right,
        axis y line*=right,
        axis x line=none,
        ylabel={LPIPS},
        scale only axis,
        height=\h,
        width=\w,
        at={(\sh,0cm)},
        ylabel shift=-6pt,
        xlabel shift=-4pt,
        ytick={0.03,0.0825,...,0.24},
        yticklabels={0.03,0.082,...,0.24},
        ymin=0.03,
        ymax=0.24,
    ]
        \pgfplotstableread[col sep=tab, colnames from=colnames]{plots/data_speed_synt.txt}\normal
        \addplot[thick, dashdotted, color=GoogleYellow, mark=+, select coords between index={0}{4}] 
        table [x index=1, y index=3] {\normal};
        \addplot[thick, dashdotted, color=GoogleBlue, mark=+, select coords between index={5}{9}] 
        table [x index=1, y index=3] {\normal};
        \addplot[thick, dashdotted, color=GoogleRed, mark=+, select coords between index={10}{14}] 
        table [x index=1, y index=3] {\normal};
    \end{axis}

\end{tikzpicture}
    \label{fig:figure1}
  \end{minipage}%
  \begin{minipage}{0.7\textwidth}
    \centering
    \includegraphics[width=\linewidth]{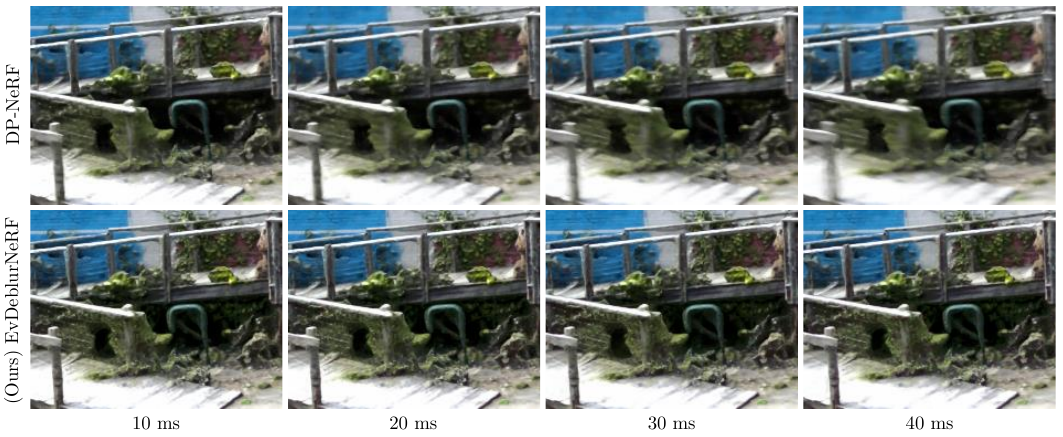}
    \label{fig:figure2}
  \end{minipage}
  \vspace{-10pt}
  \caption{Robustness to motion blur analysis on the \emph{factory} sample of Ev-DeblurBlender (left). Figures on the right show a qualitative comparison between DP-NeRF and Ev-DeblurNeRF among different exposures.}
  \label{fig:synt_speed}
\end{figure*}
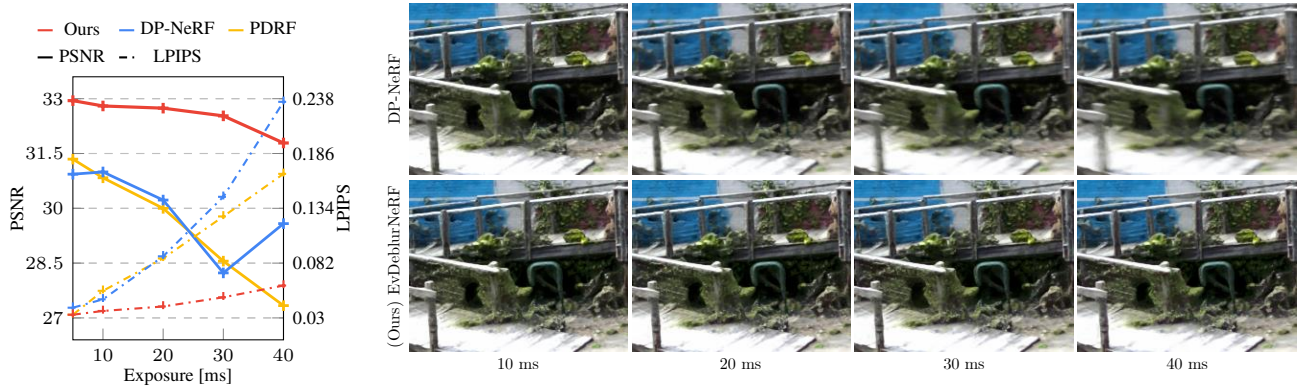

\myparagraph{Effect of using eCRF.} In Figure \ref{fig:qualitative_ablation} of this document, we complement the ablation study reported in Table 3 of the paper with a qualitative assessment of our network's key components. As discussed in the previous paragraph, the image-only architecture struggles in consistent blur conditions. Notably, incorporating event supervision significantly aids in the recovery of sharp details, as evident when comparing the first two settings in Figure \ref{fig:qualitative_ablation}. The performance further increases when adding the proposed eCRF module, as can be noticed in the checkerboard patterns on the background, the globe in the foreground, and the facial details of the figures. However, as discussed in the main paper, this improvement comes at the cost of over-augmented details and increased contrast, which are not present in the ground truth reference images. We attribute this phenomenon to the under-constrained optimization setting, which allows the eCRF module to freely augment these details as long as they appear correct once blurred though $\mathcal{L}_\text{blur}$. We solve this issue by adding an additional prior, in the form of $\mathcal{L}_\text{EDI}$, which further constrains the network in reconstructing accurate details. The improved quality is clearly demonstrated in Figure \ref{fig:qualitative_ablation}, where over-augmented details are removed, but without compromising essential details.

\myparagraph{Event-by-event vs. Event-window loss.} In this section, we extend the analysis of the robustness to training views reported in Figure 4-left of the paper. Specifically, utilizing our Ev-DeblurNeRF network, we examine the impact of implementing event supervision on an event-by-event basis, as we suggest in the paper and proposed in \cite{klenk2022nerf,low2023robust}, in contrast to accumulating events occurring over temporal windows \cite{rudnev2022eventnerf,hwang2023ev}, as well as applying supervision only at specific times during the exposure time, as in E$^2$NeRF \cite{qi2023e2nerf}. Results are reported in Figure \ref{fig:real_evbyev_vs_evbatch}. As the supervision frequency decreases, especially in sparse training views regimes, the performance also decreases. This observation aligns with the findings in \cite{low2023robust}, which suggest that noise effects and threshold variations in the event stream amplify with event accumulation, ultimately leading to a decrease in overall performance. Moreover, when only a few images are available for training, leveraging the continuous event stream to propagate absolute brightness measurements across unseen image views proves crucial for achieving top performance. Leveraging event-by-event supervision and incorporating a learnable camera response function to mitigate noise effects, our approach achieves the best performance compared to other solutions.

\myparagraph{COLMAP poses on real scenes.}
Experiments on Ev-DeblurCDAVIS presented in the paper make use of the poses obtained from the motor encoder, which tracks the camera's movement along the slider. However, in a typical real-world setting, access to such precise camera poses may not be available, although they are required by our method to work. In this section, we investigate a more general scenario where training poses are estimated using COLMAP instead of relying on the motor encoder.

Inspired by \cite{qi2023e2nerf}, we deblur training images using the EDI in Eq. (4) of the paper and then use COLMAP to estimate their poses. Analogous to the experiments conducted in the paper, we use spherical linear interpolation of the COLMAP poses to obtain poses at events' timestamps during training. At test time, we obtain the test poses by aligning the ground truth trajectory with that estimated with COLMAP. Since the two trajectories might not perfectly align, we further refine the alignment via gradient descent before computing metrics, as done in BAD-NeRF \cite{wang2022bad}, to ensure pixel-perfect aligned test poses. We also include results of our method trained on encoder poses but evaluated using refined test poses. Results are presented in Table \ref{tab:suppl_real_colmap}. 
Our method using COLMAP poses yields results comparable to those obtained using motor encoder poses, thus proving its potential in scenarios where accurate poses are not available. While performance degradation may occur in scenarios with more complex motion than that found in the Ev-DeblurCDAVIS dataset, further investigation into this aspect is left for future research endeavors.

\begin{figure*}[h!]
    \centering
    \includegraphics{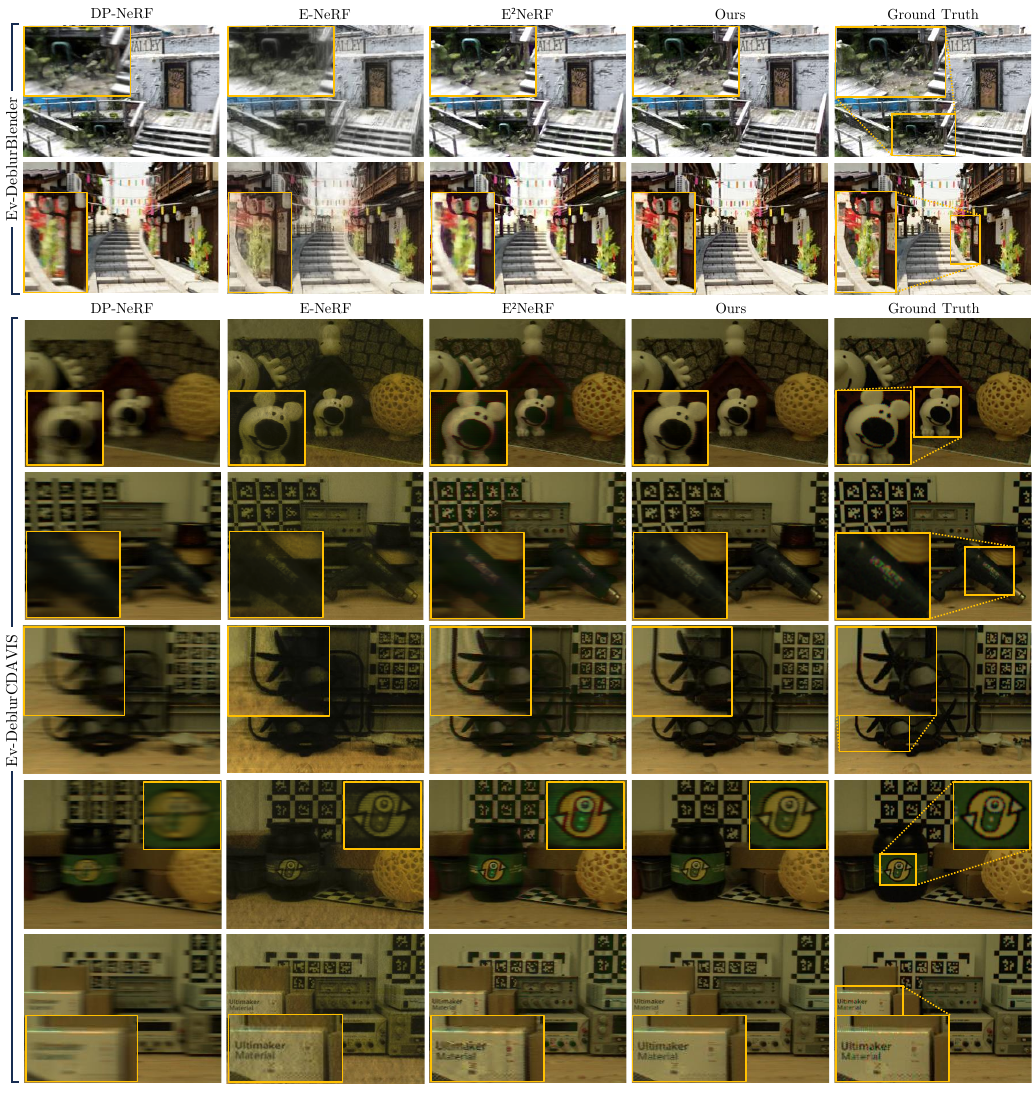}
    \caption{Qualitative comparison of synthetic (top) and real (bottom) motion blur. The remaining EV-DeblurBlender samples are provided in Figure 3 of the paper. Notice that, while the reference images can exhibit demosaicing artifacts around sharp edges, our reconstructs are less affected by these as we exploit multi-view supervision through NeRF and directly use color events without interpolation.}
    \label{fig:extended_qualitative}
\end{figure*}

\section{Additional Results} \label{sec:blur_robustness}

\paragraph{Robustness to blur.} In Figure \ref{fig:synt_speed} of this document, we supplement the analysis in Figure 4 of the main paper by comparing our Ev-DeblurNeRF network against the top-performing image-based baselines under different blur. We utilize the \emph{factory} sample of the EV-DeblurBlender dataset for this analysis since it allows us to easily control the blur intensity, and it does not constitute a corner case for the image-only baselines. We change the exposure time $\tau$ of the simulated camera in the range $\tau \in \{5, 10, 20, 30, 40\}$, which results in an average pixel displacement of $\{3, 5, 11, 16, 20\}$, and a maximum displacement of $\{15,24,50,75,96\}$ in each configuration, respectively. The quantitative and qualitative comparison in Figure \ref{fig:synt_speed} shows that using events not only helps in cases of extreme motion but also helps when the motion is not extreme. While image-only baselines recover details blindly, by trying to estimate the blur formation through a limited set of camera poses, our network can achieve higher-quality results as it exploits blur-free information carried by events at microseconds resolution. Notably, our solution shows great robustness to motion blur, while image-only performance decreases significantly as the blur increases. While these results consider synthetic data, where the effect of noise and non-idealities is limited, they underscore the promise of event cameras as complementary sensors for attaining high-quality image synthesis even in non-ideal conditions.

\myparagraph{Robustness to model mismatches.} In this section, we analyze the proposed $eCRF$ module in terms of increased robustness to model mismatches. We do so in a real setup, i.e., on the \emph{figures} sample of the Ev-DeblurCDAVIS dataset, by analyzing the sensitivity of our network to the $\Theta$ event-camera threshold. While in the paper we select $\Theta$ via manual inspection, i.e., by utilizing the event double integral \cite{pan2019bringing} as visual feedback following \cite{pan2019bringing}, we evaluate here the performance of our model when the $\Theta$ used in $\mathcal{L}_\text{ev}$ deviates from this value. We compare our network against a configuration that does not use the proposed $eCRF$, as well as a network where we also remove $\mathcal{L}_\text{EDI}$. Results are reported in Figure \ref{fig:real_threshold}. By acting in between the rendered color space and the $\mathcal{L}_\text{ev}$, $eCRF$ can modulate the brightness, or color, $\hat{L}^t$ used for computing the loss, acting as a residual between the model-based supervision (Equation (1) of the paper) and the brightness actually perceived. As a result, our solution achieves increased consistency across different choices of $\Theta$, showcasing its ability to deviate from the model-based solution in case needed.

\section{Qualitative Results and Video} \label{sec:video}
We conclude this supplementary material by including extended qualitative results. In Figure \ref{fig:extended_qualitative}, we complement Figure 3 of the paper by comparing the proposed method against top-performing networks across all remaining samples on the Ev-DeblurBlender and Ev-DeblurCDAVIS. Additionally, we provide a supplementary video showing qualitative results on all the samples of our proposed datasets. We strongly advise readers to watch our additional video where our method outperforms all baselines in rendering a novel-view continuous path, showing increased image quality and fewer artifacts than the other methods.

{
    \small
    \bibliographystyle{ieeenat_fullname}
    \bibliography{main}

\begin{thebibliography}{60}
\providecommand{\natexlab}[1]{#1}
\providecommand{\url}[1]{\texttt{#1}}
\expandafter\ifx\csname urlstyle\endcsname\relax
  \providecommand{\doi}[1]{doi: #1}\else
  \providecommand{\doi}{doi: \begingroup \urlstyle{rm}\Url}\fi

\bibitem[Azinovic et~al.(2022)Azinovic, Martin-Brualla, Goldman, Niebner, and Thies]{azinovic2022neural}
Dejan Azinovic, Ricardo Martin-Brualla, Dan~B Goldman, Matthias Niebner, and Justus Thies.
\newblock Neural {RGB}-d surface reconstruction.
\newblock In \emph{2022 {IEEE}/{CVF} Conference on Computer Vision and Pattern Recognition ({CVPR})}, pages 6290--6301. {IEEE}, 2022.

\bibitem[Barron et~al.(2021)Barron, Mildenhall, Tancik, Hedman, Martin-Brualla, and Srinivasan]{barron2021mip}
Jonathan~T Barron, Ben Mildenhall, Matthew Tancik, Peter Hedman, Ricardo Martin-Brualla, and Pratul~P Srinivasan.
\newblock Mip-{NeRF}: A multiscale representation for anti-aliasing neural radiance fields.
\newblock In \emph{2021 {IEEE}/{CVF} International Conference on Computer Vision ({ICCV})}, pages 5855--5864. {IEEE}, 2021.

\bibitem[Barron et~al.(2022)Barron, Mildenhall, Verbin, Srinivasan, and Hedman]{barron2022mip}
Jonathan~T Barron, Ben Mildenhall, Dor Verbin, Pratul~P Srinivasan, and Peter Hedman.
\newblock Mip-{NeRF} 360: Unbounded anti-aliased neural radiance fields.
\newblock In \emph{2022 {IEEE}/{CVF} Conference on Computer Vision and Pattern Recognition ({CVPR})}, pages 5470--5479. {IEEE}, 2022.

\bibitem[Bhattacharya et~al.(2023)Bhattacharya, Madaan, Cladera, Vemprala, Bonatti, Daniilidis, Kapoor, Kumar, Matni, and Gupta]{bhattacharya2023evdnerf}
Anish Bhattacharya, Ratnesh Madaan, Fernando Cladera, Sai Vemprala, Rogerio Bonatti, Kostas Daniilidis, Ashish Kapoor, Vijay Kumar, Nikolai Matni, and Jayesh~K Gupta.
\newblock Evdnerf: Reconstructing event data with dynamic neural radiance fields.
\newblock \emph{arXiv preprint arXiv:2310.02437}, 2023.

\bibitem[Chen et~al.(2022)Chen, Xu, Geiger, Yu, and Su]{chen2022tensorf}
Anpei Chen, Zexiang Xu, Andreas Geiger, Jingyi Yu, and Hao Su.
\newblock {TensoRF}: Tensorial radiance fields.
\newblock In \emph{Lecture Notes in Computer Science}, pages 333--350. Springer Nature Switzerland, 2022.

\bibitem[Cheng and Rama(2023)]{peng2022pdrf}
Peng Cheng and Chellappa Rama.
\newblock Pdrf: Progressively deblurring radiance field for fast and robust scene reconstruction from blurry images, 2023.

\bibitem[Gallego et~al.(2022)Gallego, Delbruck, Orchard, Bartolozzi, Taba, Censi, Leutenegger, Davison, Conradt, Daniilidis, and Scaramuzza]{gallego2022survey}
Guillermo Gallego, Tobi Delbruck, Garrick Orchard, Chiara Bartolozzi, Brian Taba, Andrea Censi, Stefan Leutenegger, Andrew~J. Davison, Jorg Conradt, Kostas Daniilidis, and Davide Scaramuzza.
\newblock Event-based vision: A survey.
\newblock \emph{{IEEE} Transactions on Pattern Analysis and Machine Intelligence}, 44\penalty0 (1):\penalty0 154--180, 2022.

\bibitem[Gao et~al.(2022)Gao, Gao, He, Lu, Xu, and Li]{gao2022nerf}
Kyle Gao, Yina Gao, Hongjie He, Denning Lu, Linlin Xu, and Jonathan Li.
\newblock Nerf: Neural radiance field in 3d vision, a comprehensive review.
\newblock \emph{arXiv preprint arXiv:2210.00379}, 2022.

\bibitem[Han et~al.(2020)Han, Zhou, Duan, Tang, Xu, Xu, Huang, and Shi]{han2020neuromorphic}
Jin Han, Chu Zhou, Peiqi Duan, Yehui Tang, Chang Xu, Chao Xu, Tiejun Huang, and Boxin Shi.
\newblock Neuromorphic camera guided high dynamic range imaging.
\newblock In \emph{2020 {IEEE}/{CVF} Conference on Computer Vision and Pattern Recognition ({CVPR})}, pages 1730--1739. {IEEE}, 2020.

\bibitem[Haoyu et~al.(2020)Haoyu, Minggui, Boxin, YIzhou, and Tiejun]{haoyu2020learning}
Chen Haoyu, Teng Minggui, Shi Boxin, Wang YIzhou, and Huang Tiejun.
\newblock Learning to deblur and generate high frame rate video with an event camera.
\newblock \emph{arXiv preprint arXiv:2003.00847}, 2020.

\bibitem[Hu et~al.(2021)Hu, Liu, and Delbruck]{hu2021v2e}
Yuhuang Hu, Shih-Chii Liu, and Tobi Delbruck.
\newblock v2e: From video frames to realistic {DVS} events.
\newblock In \emph{2021 {IEEE}/{CVF} Conference on Computer Vision and Pattern Recognition Workshops ({CVPRW})}, pages 1312--1321. {IEEE}, 2021.

\bibitem[Huang et~al.(2022)Huang, Zhang, Feng, Li, Wang, and Wang]{huang2022hdr}
Xin Huang, Qi Zhang, Ying Feng, Hongdong Li, Xuan Wang, and Qing Wang.
\newblock {HDR}-{NeRF}: High dynamic range neural radiance fields.
\newblock In \emph{2022 {IEEE}/{CVF} Conference on Computer Vision and Pattern Recognition ({CVPR})}, pages 18398--18408. {IEEE}, 2022.

\bibitem[Hwang et~al.(2023)Hwang, Kim, and Kim]{hwang2023ev}
Inwoo Hwang, Junho Kim, and Young~Min Kim.
\newblock Ev-{NeRF}: Event based neural radiance field.
\newblock In \emph{2023 {IEEE}/{CVF} Winter Conference on Applications of Computer Vision ({WACV})}, pages 837--847. {IEEE}, 2023.

\bibitem[Jiang et~al.(2020)Jiang, Zhang, Zou, Ren, Lv, and Liu]{jiang2020learning}
Zhe Jiang, Yu Zhang, Dongqing Zou, Jimmy Ren, Jiancheng Lv, and Yebin Liu.
\newblock Learning event-based motion deblurring.
\newblock In \emph{2020 {IEEE}/{CVF} Conference on Computer Vision and Pattern Recognition ({CVPR})}, pages 3320--3329. {IEEE}, 2020.

\bibitem[Kingma and Ba(2014)]{kingma2014adam}
Diederik~P Kingma and Jimmy Ba.
\newblock Adam: A method for stochastic optimization.
\newblock \emph{arXiv preprint arXiv:1412.6980}, 2014.

\bibitem[Klenk et~al.(2023)Klenk, Koestler, Scaramuzza, and Cremers]{klenk2022nerf}
Simon Klenk, Lukas Koestler, Davide Scaramuzza, and Daniel Cremers.
\newblock E-{NeRF}: Neural radiance fields from a moving event camera.
\newblock \emph{{IEEE} Robotics and Automation Letters}, 8\penalty0 (3):\penalty0 1587--1594, 2023.

\bibitem[Kundu et~al.(2022)Kundu, Genova, Yin, Fathi, Pantofaru, Guibas, Tagliasacchi, Dellaert, and Funkhouser]{kundu2022panoptic}
Abhijit Kundu, Kyle Genova, Xiaoqi Yin, Alireza Fathi, Caroline Pantofaru, Leonidas Guibas, Andrea Tagliasacchi, Frank Dellaert, and Thomas Funkhouser.
\newblock Panoptic neural fields: A semantic object-aware neural scene representation.
\newblock In \emph{2022 {IEEE}/{CVF} Conference on Computer Vision and Pattern Recognition ({CVPR})}, pages 12871--12881. {IEEE}, 2022.

\bibitem[Lee et~al.(2023)Lee, Lee, Shin, and Lee]{lee2023dp}
Dogyoon Lee, Minhyeok Lee, Chajin Shin, and Sangyoun Lee.
\newblock {DP}-{NeRF}: Deblurred neural radiance field with physical scene priors.
\newblock In \emph{2023 {IEEE}/{CVF} Conference on Computer Vision and Pattern Recognition ({CVPR})}, pages 12386--12396. {IEEE}, 2023.

\bibitem[Li et~al.(2015)Li, Brandli, Berner, Liu, Yang, Liu, and Delbruck]{li2015design}
Chenghan Li, Christian Brandli, Raphael Berner, Hongjie Liu, Minhao Yang, Shih-Chii Liu, and Tobi Delbruck.
\newblock Design of an {RGBW} color {VGA} rolling and global shutter dynamic and active-pixel vision sensor.
\newblock In \emph{2015 {IEEE} International Symposium on Circuits and Systems ({ISCAS})}, pages 718--721. IEEE, {IEEE}, 2015.

\bibitem[Li et~al.(2021)Li, Niklaus, Snavely, and Wang]{li2021neural}
Zhengqi Li, Simon Niklaus, Noah Snavely, and Oliver Wang.
\newblock Neural scene flow fields for space-time view synthesis of dynamic scenes.
\newblock In \emph{2021 {IEEE}/{CVF} Conference on Computer Vision and Pattern Recognition ({CVPR})}, pages 6498--6508. {IEEE}, 2021.

\bibitem[Lin et~al.(2021)Lin, Ma, Torralba, and Lucey]{lin2021barf}
Chen-Hsuan Lin, Wei-Chiu Ma, Antonio Torralba, and Simon Lucey.
\newblock {BARF}: Bundle-adjusting neural radiance fields.
\newblock In \emph{2021 {IEEE}/{CVF} International Conference on Computer Vision ({ICCV})}, pages 5741--5751. {IEEE}, 2021.

\bibitem[Liu et~al.(2023)Liu, Milano, Frey, Siegwart, Blum, and Cadena]{liu2022unsupervised}
Zhizheng Liu, Francesco Milano, Jonas Frey, Roland Siegwart, Hermann Blum, and Cesar Cadena.
\newblock Unsupervised continual semantic adaptation through neural rendering.
\newblock In \emph{Proceedings of the IEEE/CVF Conference on Computer Vision and Pattern Recognition}, pages 3031--3040, 2023.

\bibitem[Low and Lee(2023)]{low2023robust}
Weng~Fei Low and Gim~Hee Lee.
\newblock Robust e-nerf: Nerf from sparse \& noisy events under non-uniform motion.
\newblock In \emph{Proceedings of the IEEE/CVF International Conference on Computer Vision}, pages 18335--18346, 2023.

\bibitem[Ma et~al.(2022)Ma, Li, Liao, Zhang, Wang, Wang, and Sander]{ma2022deblur}
Li Ma, Xiaoyu Li, Jing Liao, Qi Zhang, Xuan Wang, Jue Wang, and Pedro~V Sander.
\newblock Deblur-{NeRF}: Neural radiance fields from blurry images.
\newblock In \emph{2022 {IEEE}/{CVF} Conference on Computer Vision and Pattern Recognition ({CVPR})}, pages 12861--12870. {IEEE}, 2022.

\bibitem[Messikommer et~al.(2022)Messikommer, Georgoulis, Gehrig, Tulyakov, Erbach, Bochicchio, Li, and Scaramuzza]{messikommer2022multi}
Nico Messikommer, Stamatios Georgoulis, Daniel Gehrig, Stepan Tulyakov, Julius Erbach, Alfredo Bochicchio, Yuanyou Li, and Davide Scaramuzza.
\newblock Multi-bracket high dynamic range imaging with event cameras.
\newblock In \emph{2022 {IEEE}/{CVF} Conference on Computer Vision and Pattern Recognition Workshops ({CVPRW})}, pages 547--557. {IEEE}, 2022.

\bibitem[Messikommer et~al.(2023)Messikommer, Fang, Gehrig, and Scaramuzza]{messikommer2023data}
Nico Messikommer, Carter Fang, Mathias Gehrig, and Davide Scaramuzza.
\newblock Data-driven feature tracking for event cameras.
\newblock In \emph{2023 {IEEE}/{CVF} Conference on Computer Vision and Pattern Recognition ({CVPR})}, pages 5642--5651. {IEEE}, 2023.

\bibitem[Mildenhall et~al.(2020)Mildenhall, Srinivasan, Tancik, Barron, Ramamoorthi, and Ng]{mildenhall2020nerf}
Ben Mildenhall, Pratul~P. Srinivasan, Matthew Tancik, Jonathan~T. Barron, Ravi Ramamoorthi, and Ren Ng.
\newblock Nerf: Representing scenes as neural radiance fields for view synthesis.
\newblock In \emph{ECCV}, 2020.

\bibitem[Mildenhall et~al.(2022)Mildenhall, Hedman, Martin-Brualla, Srinivasan, and Barron]{mildenhall2022nerf}
Ben Mildenhall, Peter Hedman, Ricardo Martin-Brualla, Pratul~P Srinivasan, and Jonathan~T Barron.
\newblock {NeRF} in the dark: High dynamic range view synthesis from noisy raw images.
\newblock In \emph{2022 {IEEE}/{CVF} Conference on Computer Vision and Pattern Recognition ({CVPR})}, pages 16190--16199. {IEEE}, 2022.

\bibitem[Pan et~al.(2019)Pan, Scheerlinck, Yu, Hartley, Liu, and Dai]{pan2019bringing}
Liyuan Pan, Cedric Scheerlinck, Xin Yu, Richard Hartley, Miaomiao Liu, and Yuchao Dai.
\newblock Bringing a blurry frame alive at high frame-rate with an event camera.
\newblock In \emph{2019 {IEEE}/{CVF} Conference on Computer Vision and Pattern Recognition ({CVPR})}, pages 6820--6829. {IEEE}, 2019.

\bibitem[Paszke et~al.(2019)Paszke, Gross, Massa, Lerer, Bradbury, Chanan, Killeen, Lin, Gimelshein, Antiga, Desmaison, Kopf, Yang, DeVito, Raison, Tejani, Chilamkurthy, Steiner, Fang, Bai, and Chintala]{paszke2019pytorch}
Adam Paszke, Sam Gross, Francisco Massa, Adam Lerer, James Bradbury, Gregory Chanan, Trevor Killeen, Zeming Lin, Natalia Gimelshein, Luca Antiga, Alban Desmaison, Andreas Kopf, Edward Yang, Zachary DeVito, Martin Raison, Alykhan Tejani, Sasank Chilamkurthy, Benoit Steiner, Lu Fang, Junjie Bai, and Soumith Chintala.
\newblock Pytorch: An imperative style, high-performance deep learning library.
\newblock In \emph{Advances in Neural Information Processing Systems 32}, pages 8024--8035. Curran Associates, Inc., 2019.

\bibitem[Qi et~al.(2023)Qi, Zhu, Zhang, and Li]{qi2023e2nerf}
Yunshan Qi, Lin Zhu, Yu Zhang, and Jia Li.
\newblock E2nerf: Event enhanced neural radiance fields from blurry images.
\newblock In \emph{Proceedings of the IEEE/CVF International Conference on Computer Vision}, pages 13254--13264, 2023.

\bibitem[Rebecq et~al.(2018)Rebecq, Gehrig, and Scaramuzza]{rebecq2018esim}
Henri Rebecq, Daniel Gehrig, and Davide Scaramuzza.
\newblock Esim: an open event camera simulator.
\newblock In \emph{2nd Annual Conference on Robot Learning, CoRL 2018, Z{\"{u}}rich, Switzerland, 29-31 October 2018, Proceedings}, pages 969--982. {PMLR}, 2018.

\bibitem[Rudnev et~al.(2023)Rudnev, Elgharib, Theobalt, and Golyanik]{rudnev2022eventnerf}
Viktor Rudnev, Mohamed Elgharib, Christian Theobalt, and Vladislav Golyanik.
\newblock Eventnerf: Neural radiance fields from a single colour event camera.
\newblock In \emph{IEEE Conf. Comput. Vis. Pattern Recog.}, 2023.

\bibitem[Shang et~al.(2021)Shang, Ren, Zou, Ren, Luo, and Zuo]{shang2021bringing}
Wei Shang, Dongwei Ren, Dongqing Zou, Jimmy~S Ren, Ping Luo, and Wangmeng Zuo.
\newblock Bringing events into video deblurring with non-consecutively blurry frames.
\newblock In \emph{2021 {IEEE}/{CVF} International Conference on Computer Vision ({ICCV})}, pages 4531--4540. {IEEE}, 2021.

\bibitem[Shoemake(1985)]{shoemake1985animating}
Ken Shoemake.
\newblock Animating rotation with quaternion curves.
\newblock \emph{{ACM} {SIGGRAPH} Computer Graphics}, 19\penalty0 (3):\penalty0 245--254, 1985.

\bibitem[Son et~al.(2021)Son, Lee, Lee, Cho, and Lee]{son2021recurrent}
Hyeongseok Son, Junyong Lee, Jonghyeop Lee, Sunghyun Cho, and Seungyong Lee.
\newblock Recurrent video deblurring with blur-invariant motion estimation and pixel volumes.
\newblock \emph{ACM Transactions on Graphics}, 40\penalty0 (5):\penalty0 1--18, 2021.

\bibitem[Sucar et~al.(2021)Sucar, Liu, Ortiz, and Davison]{sucar2021implicit}
Edgar Sucar, Shikun Liu, Joseph Ortiz, and Andrew~J. Davison.
\newblock \emph{{iMAP}: Implicit Mapping and Positioning in Real-Time}, pages 6229--6238.
\newblock {IEEE}, 2021.

\bibitem[Sun et~al.(2022)Sun, Sakaridis, Liang, Jiang, Yang, Sun, Ye, Wang, and Gool]{sun2022event}
Lei Sun, Christos Sakaridis, Jingyun Liang, Qi Jiang, Kailun Yang, Peng Sun, Yaozu Ye, Kaiwei Wang, and Luc~Van Gool.
\newblock Event-based fusion for~motion deblurring with~cross-modal attention.
\newblock In \emph{Lecture Notes in Computer Science}, pages 412--428. Springer, Springer Nature Switzerland, 2022.

\bibitem[Sun et~al.(2023)Sun, Sakaridis, Liang, Sun, Cao, Zhang, Jiang, Wang, and Van~Gool]{sun2023event}
Lei Sun, Christos Sakaridis, Jingyun Liang, Peng Sun, Jiezhang Cao, Kai Zhang, Qi Jiang, Kaiwei Wang, and Luc Van~Gool.
\newblock Event-based frame interpolation with ad-hoc deblurring.
\newblock In \emph{Proceedings of the IEEE/CVF Conference on Computer Vision and Pattern Recognition}, pages 18043--18052, 2023.

\bibitem[Tewari et~al.(2022)Tewari, Thies, Mildenhall, Srinivasan, Tretschk, Yifan, Lassner, Sitzmann, Martin-Brualla, Lombardi, et~al.]{tewari2022advances}
Ayush Tewari, Justus Thies, Ben Mildenhall, Pratul Srinivasan, Edgar Tretschk, Wang Yifan, Christoph Lassner, Vincent Sitzmann, Ricardo Martin-Brualla, Stephen Lombardi, et~al.
\newblock Advances in neural rendering.
\newblock In \emph{Computer Graphics Forum}, pages 703--735. Wiley Online Library, 2022.

\bibitem[Tulyakov et~al.(2021)Tulyakov, Gehrig, Georgoulis, Erbach, Gehrig, Li, and Scaramuzza]{tulyakov2021time}
Stepan Tulyakov, Daniel Gehrig, Stamatios Georgoulis, Julius Erbach, Mathias Gehrig, Yuanyou Li, and Davide Scaramuzza.
\newblock Time lens: Event-based video frame interpolation.
\newblock In \emph{2021 {IEEE}/{CVF} Conference on Computer Vision and Pattern Recognition ({CVPR})}, pages 16155--16164. {IEEE}, 2021.

\bibitem[Tulyakov et~al.(2022)Tulyakov, Bochicchio, Gehrig, Georgoulis, Li, and Scaramuzza]{tulyakov2022time}
Stepan Tulyakov, Alfredo Bochicchio, Daniel Gehrig, Stamatios Georgoulis, Yuanyou Li, and Davide Scaramuzza.
\newblock Time lens++: Event-based frame interpolation with parametric nonlinear flow and multi-scale fusion.
\newblock In \emph{2022 {IEEE}/{CVF} Conference on Computer Vision and Pattern Recognition ({CVPR})}, pages 17755--17764. {IEEE}, 2022.

\bibitem[Union(2011)]{bt2011studio}
International~Telecomunication Union.
\newblock Studio encoding parameters of digital television for standard 4: 3 and wide-screen 16: 9 aspect ratios. bt.709.
\newblock \emph{Technical Report, International Telecomunication Union}, 2011.

\bibitem[Verbin et~al.(2022)Verbin, Hedman, Mildenhall, Zickler, Barron, and Srinivasan]{verbin2022ref}
Dor Verbin, Peter Hedman, Ben Mildenhall, Todd Zickler, Jonathan~T Barron, and Pratul~P Srinivasan.
\newblock Ref-{NeRF}: Structured view-dependent appearance for neural radiance fields.
\newblock In \emph{2022 {IEEE}/{CVF} Conference on Computer Vision and Pattern Recognition ({CVPR})}, pages 5481--5490. IEEE, {IEEE}, 2022.

\bibitem[Vidal et~al.(2018)Vidal, Rebecq, Horstschaefer, and Scaramuzza]{vidal2018ultimate}
Antoni~Rosinol Vidal, Henri Rebecq, Timo Horstschaefer, and Davide Scaramuzza.
\newblock Ultimate slam? combining events, images, and imu for robust visual slam in hdr and high-speed scenarios.
\newblock \emph{IEEE Robotics and Automation Letters}, 3\penalty0 (2):\penalty0 994--1001, 2018.

\bibitem[Vitoria et~al.(2022)Vitoria, Georgoulis, Tulyakov, Bochicchio, Erbach, and Li]{vitoria2023event}
Patricia Vitoria, Stamatios Georgoulis, Stepan Tulyakov, Alfredo Bochicchio, Julius Erbach, and Yuanyou Li.
\newblock Event-based image deblurring with~dynamic motion awareness.
\newblock In \emph{Eur. Conf. Comput. Vis.} Springer Nature Switzerland, 2022.

\bibitem[Wang et~al.(2020)Wang, He, Yu, Xia, and Yang]{wang2020event}
Bishan Wang, Jingwei He, Lei Yu, Gui-Song Xia, and Wen Yang.
\newblock Event enhanced high-quality image recovery.
\newblock In \emph{Computer Vision {\textendash} {ECCV} 2020}, pages 155--171. Springer International Publishing, 2020.

\bibitem[Wang et~al.(2022)Wang, Wu, Guo, Zhang, Tai, and Hu]{wang2022nerf}
Chen Wang, Xian Wu, Yuan-Chen Guo, Song-Hai Zhang, Yu-Wing Tai, and Shi-Min Hu.
\newblock {NeRF}-{SR}: High quality neural radiance fields using supersampling.
\newblock In \emph{Proceedings of the 30th {ACM} International Conference on Multimedia}, pages 6445--6454. {ACM}, 2022.

\bibitem[Wang et~al.(2021)Wang, Liu, Liu, Theobalt, Komura, and Wang]{wang2021neus}
Peng Wang, Lingjie Liu, Yuan Liu, Christian Theobalt, Taku Komura, and Wenping Wang.
\newblock Neus: Learning neural implicit surfaces by volume rendering for multi-view reconstruction.
\newblock \emph{NeurIPS}, 2021.

\bibitem[Wang et~al.(2023)Wang, Zhao, Ma, and Liu]{wang2022bad}
Peng Wang, Lingzhe Zhao, Ruijie Ma, and Peidong Liu.
\newblock Bad-nerf: Bundle adjusted deblur neural radiance fields.
\newblock In \emph{Proceedings of the IEEE/CVF Conference on Computer Vision and Pattern Recognition}, pages 4170--4179, 2023.

\bibitem[Wu et~al.(2022)Wu, You, He, Yang, Tian, Wang, Zhang, and Liao]{wu2022video}
Song Wu, Kaichao You, Weihua He, Chen Yang, Yang Tian, Yaoyuan Wang, Ziyang Zhang, and Jianxing Liao.
\newblock Video interpolation by~event-driven anisotropic adjustment of~optical flow.
\newblock In \emph{Lecture Notes in Computer Science}, pages 267--283. Springer Nature Switzerland, 2022.

\bibitem[Xie et~al.(2021)Xie, Park, Martin-Brualla, and Brown]{xie2021fig}
Christopher Xie, Keunhong Park, Ricardo Martin-Brualla, and Matthew Brown.
\newblock {FiG}-{NeRF}: Figure-ground neural radiance fields for 3d object category modelling.
\newblock In \emph{2021 International Conference on 3D Vision (3DV)}, pages 962--971. IEEE, {IEEE}, 2021.

\bibitem[Xu et~al.(2021)Xu, Yu, Wang, Yang, Xia, Jia, Qiao, and Liu]{xu2021motion}
Fang Xu, Lei Yu, Bishan Wang, Wen Yang, Gui-Song Xia, Xu Jia, Zhendong Qiao, and Jianzhuang Liu.
\newblock Motion deblurring with real events.
\newblock In \emph{2021 {IEEE}/{CVF} International Conference on Computer Vision ({ICCV})}, pages 2583--2592. {IEEE}, 2021.

\bibitem[Yen-Chen et~al.(2021)Yen-Chen, Florence, Barron, Rodriguez, Isola, and Lin]{yen2021inerf}
Lin Yen-Chen, Pete Florence, Jonathan~T Barron, Alberto Rodriguez, Phillip Isola, and Tsung-Yi Lin.
\newblock {iNeRF}: Inverting neural radiance fields for pose estimation.
\newblock In \emph{2021 {IEEE}/{RSJ} International Conference on Intelligent Robots and Systems ({IROS})}, pages 1323--1330. IEEE, {IEEE}, 2021.

\bibitem[Yu et~al.(2022)Yu, Peng, Niemeyer, Sattler, and Geiger]{yu2022monosdf}
Zehao Yu, Songyou Peng, Michael Niemeyer, Torsten Sattler, and Andreas Geiger.
\newblock Monosdf: Exploring monocular geometric cues for neural implicit surface reconstruction.
\newblock \emph{Advances in Neural Information Processing Systems (NeurIPS)}, 2022.

\bibitem[Zamir et~al.(2021)Zamir, Arora, Khan, Hayat, Khan, Yang, and Shao]{zamir2021multi}
Syed~Waqas Zamir, Aditya Arora, Salman Khan, Munawar Hayat, Fahad~Shahbaz Khan, Ming-Hsuan Yang, and Ling Shao.
\newblock Multi-stage progressive image restoration.
\newblock In \emph{2021 IEEE/CVF Conference on Computer Vision and Pattern Recognition (CVPR)}, pages 14821--14831. IEEE, 2021.

\bibitem[Zhang et~al.(2022)Zhang, Zhang, Dai, Li, and Koniusz]{zhang2023event}
Hongguang Zhang, Limeng Zhang, Yuchao Dai, Hongdong Li, and Piotr Koniusz.
\newblock Event-guided multi-patch network with self-supervision for non-uniform motion deblurring.
\newblock \emph{International Journal of Computer Vision}, 131\penalty0 (2):\penalty0 453--470, 2022.

\bibitem[Zhang et~al.(2020)Zhang, Zhang, Chen, and Wang]{zhang2020hybrid}
Limeng Zhang, Hongguang Zhang, Jihua Chen, and Lei Wang.
\newblock Hybrid deblur net: Deep non-uniform deblurring with event camera.
\newblock \emph{{IEEE} Access}, 8:\penalty0 148075--148083, 2020.

\bibitem[Zhang et~al.(2018)Zhang, Isola, Efros, Shechtman, and Wang]{zhang2018unreasonable}
Richard Zhang, Phillip Isola, Alexei~A Efros, Eli Shechtman, and Oliver Wang.
\newblock The unreasonable effectiveness of deep features as a perceptual metric.
\newblock In \emph{2018 IEEE/CVF Conference on Computer Vision and Pattern Recognition}, pages 586--595. IEEE, 2018.

\bibitem[Zhu et~al.(2022)Zhu, Peng, Larsson, Xu, Bao, Cui, Oswald, and Pollefeys]{zhu2022neural}
Zihan Zhu, Songyou Peng, Viktor Larsson, Weiwei Xu, Hujun Bao, Zhaopeng Cui, Martin~R. Oswald, and Marc Pollefeys.
\newblock {NICE}-{SLAM}: Neural implicit scalable encoding for {SLAM}.
\newblock In \emph{2022 {IEEE}/{CVF} Conference on Computer Vision and Pattern Recognition ({CVPR})}, pages 12--786. {IEEE}, 2022.

\end{thebibliography}
}

\end{document}